## TITLE PAGE

**Title: Rough set based lattice structure for knowledge representation in medical expert systems: low back pain management case study**

- **Author Names and Affiliations**

Debarpita Santra[a], Swapan Kumar Basu[b], Jyotsna Kumar Mandal[c], Subrata Goswami[d]

[a,c]Department of Computer Science and Engineering, Faculty of Engineering, Technology and Management, University of Kalyani, Block C, Nadia, Kalyani, West Bengal, PIN - 741245, India

[b]Department of Computer Science, Institute of Science, Banaras Hindu University, Varanasi-221005, Uttar Pradesh, India

[d]ESI Institute of Pain Management, ESI Hospital Sealdah premises, 301/3 Acharya Prafulla Chandra Road, Kolkata – 700009, West Bengal, India

- **Contact Details of Authors:**

**1st Author Name:** Debarpita Santra
**Email Address:** debarpita.cs@gmail.com
**Telephone Number:** +91-8961468489
**ORCID of the author:** 0000-0001-5651-5642
**Full postal address:** Department of Computer Science and Engineering, Faculty of Engineering, Technology and Management, University of Kalyani, Block C, Nadia, Kalyani, West Bengal, PIN - 741245, India

**2nd Author Name:** Swapan Kumar Basu
**Email Address:** swapankb@gmail.com
**Telephone Number:** +91-9450541334
**Full postal address:** Department of Computer Science, Institute of Science, Banaras Hindu University, Varanasi-221005, Uttar Pradesh, India

**3rd Author Name:** Jyotsna Kumar Mandal
**Email Address:** jkm.cse@gmail.com
**Telephone Number:** +91-9434352214
**Full postal address:** Department of Computer Science and Engineering, Faculty of Engineering, Technology and Management, University of Kalyani, Block C, Nadia, Kalyani, West Bengal, PIN - 741245, India

**4th Author Name:** Subrata Goswami
**Email Address:** drsgoswami@gmail.com
**Telephone Number:** +91-9830430430
**Full postal address:** ESI Institute of Pain Management, ESI Hospital Sealdah premises, 301/3 Acharya Prafulla Chandra Road, Kolkata – 700009, West Bengal, India

- **Corresponding Author**

**Name:** Debarpita Santra

**Affiliation:** Department of Computer Science and Engineering, Faculty of Engineering, Technology and Management, University of Kalyani, Block C, Nadia, Kalyani, West Bengal, PIN - 741245, India

**Email Address:** debarpita.cs@gmail.com

**Telephone Number:** +91-8961468489

**Full postal address:** Department of Computer Science and Engineering, Faculty of Engineering, Technology and Management, University of Kalyani, Block C, Nadia, Kalyani, West Bengal, PIN - 741245, India

**Abstract**

The aim of medical knowledge representation is to capture the detailed domain knowledge in a clinically efficient manner and to offer a reliable resolution with the acquired knowledge. The knowledge base to be used by a medical expert system should allow incremental growth with inclusion of updated knowledge over the time. As knowledge are gathered from a variety of knowledge sources by different knowledge engineers, the problem of redundancy is an important concern here due to increased processing time of knowledge and occupancy of large computational storage to accommodate all the gathered knowledge. Also there may exist many inconsistent knowledge in the knowledge base. In this paper, we have proposed a rough set based lattice structure for knowledge representation in medical expert systems which overcomes the problem of redundancy and inconsistency in knowledge and offers computational efficiency with respect to both time and space. We have also generated an optimal set of decision rules that would be used directly by the inference engine. The reliability of each rule has been measured using a new metric called credibility factor, and the certainty and coverage factors of a decision rule have been re-defined. With a set of decisions rules arranged in descending order according to their reliability measures, the medical expert system will consider the highly reliable and certain rules at first, then it would search for the possible and uncertain rules at later stage, if recommended by physicians. The proposed knowledge representation technique has been illustrated using an example from the domain of low back pain. The proposed scheme ensures completeness, consistency, integrity, non-redundancy, and ease of access.

## 1. Introduction

Medical diagnosis is the art of employing physician's knowledge and intuition in determining the disease of a patient. Exponential growth of medical knowledge and its incredible advancements over years are making the diagnosis process more complicated, thereby deteriorating the quality of medical care. Also, from the speed and reliability point of view, human beings have their limitations, because of excessive patient load, human errors, fatigue, and not having up-to-date knowledge about the recent advancements in medical and associated fields. A medical expert system (Shortliffe (1986); Persidis and Persidis (1991); Abu-Nasser (2017)) which is basically a computer system (devoid of human limitations) powered by AI techniques using the medical domain knowledge, can come to the help and assistance to the doctors.

Efficiency of a medical expert system depends on how much updated knowledge is stored in its knowledge base and how easily it can be retrieved. A knowledge base contains all the relevant facts and rules associated with the problem domain. Knowledge is acquired from different knowledge sources such as domain specific books, related research papers, patient case histories and wisdom gleaned from expert physicians in related medical domain. In this paper, we have proposed an efficient knowledge representation technique which ensures completeness, consistency, integrity, ease of access, and zero redundancy. The proposed knowledge representation technique automatically generates a set of optimal rules to be used directly by the inference engine of a medical expert system.

The Knowledge Principle: "*A system exhibits intelligent understanding and action at a high level of competence primarily because of the specific knowledge that it can bring to bear: the concepts, facts, representations, methods, models, metaphors, and heuristics about its domain of endeavor*." - (Lenat and Feigenbaum (1992))

Design of a knowledge base should allow incremental growth with additions of new medical knowledge over the time. Suppose a knowledge engineer, who is responsible for knowledge acquisition, interviews a domain expert and inserts the acquired knowledge in knowledge base at time $t_1$. At time instant $t_2$ ($>t_1$) the knowledge engineer gathers additional knowledge from a different knowledge source and also inserts the gathered knowledge in the knowledge base. Some other knowledge engineers may acquire knowledge about the same medical domain from some other knowledge sources. As the knowledge are acquired from a variety of sources like books, journals, research articles, patient case histories etc., the knowledge base may include some redundant and inconsistent knowledge sometimes. To overcome the problem of redundancy in the gathered knowledge, we have proposed a lattice based knowledge representation technique for constructing the knowledge base. As lattice structure (Jongsma (2016)) is purely based on set theory which does not allow duplicate elements, the knowledge base is assured to have zero redundancy. Also for handling inconsistent knowledge in the knowledge base, we have applied Rough Set theory (Pawlak et al. 2008; Pawlak (1998); Zhang et al. 2016; Zhenxiang et al. 1999) on the lattice based design. The Rough Set theory helps us generate certain, possible, and uncertain rules from the inconsistent knowledge. We have also proposed a rule minimization technique for generating optimal rules. This results in a compact knowledge base design.

As we want to acquire sufficient knowledge about a medical domain for a medical expert system in order to help the latter in reliable inferencing, we first have to identify the mutually exclusive knowledge sets for the domain under consideration. More specifically, let us consider $\chi$ as the entire set of knowledge acquired till time $t$. Suppose $\chi_1$, $\chi_2$, …, $\chi_k$ ($k > 0$) are different sets of knowledge, where $\chi_1 \cup \chi_2 \cup \ldots \cup \chi_k = \chi$ and $\chi_1 \cap \chi_2 \cap \ldots \cap \chi_k = \phi$. Every set of knowledge contains similar kinds of knowledge. The examples of different sets of knowledge include the knowledge about radicular low back pain (LBP) (Allegri et al. 2016), knowledge about localized LBP (Natvig et al. 2001), past medical history of patients, knowledge regarding special examinations for LBP (*https://stanfordmedicine25.stanford.edu/the25/BackExam.html*, accessed on August 19, 2018), and so on. If we take an arbitrary pair ($\chi_i$, $\chi_j$) [$i \neq j$] such that $\chi_i$ will not be used in inference in presence of $\chi_j$, then we say that the pair is mutually exclusive. A knowledge set may also comprise many pairs of mutually exclusive knowledge sets. For example, the knowledge about symptoms of LBP with radicular leg pain and symptoms of LBP without radiculopathy form two different mutually exclusive knowledge sets as they both cannot appear simultaneously during the resolution process.

Designing a knowledge base using a single lattice structure may become large in size. Instead, we represent mutually exclusive knowledge sets using lattice-based structure. If a set of knowledge does not have any mutually exclusive knowledge sets, we would then apply the lattice structure on the whole set. So, each lattice structure would appear as a manageable-sized knowledge module for performing various operations like insertion, deletion or modification of knowledge on it. This kind of modular lattice based approach for knowledge representation will surely enhance efficiency of inference mechanism of the medical expert system. Instead of instantiating all the knowledge modules at a time, this kind of representation would call the modules whenever needed.

The rest of the paper is organized as: section 2 gives an overview of related works. In section 3, we discuss about the basics of lattice and rough set theory. In section 4, we have discussed about our proposed rough set based lattice structure for medical knowledge representation. Section 5 illustrates the proposed knowledge representation technique using a simple example from the domain of LBP. In section 6, we discuss about the basic properties of the knowledge representation. Section 7 concludes the paper.

## 2. Previous work

Development of an effective knowledge base poses a great challenge to the expert-system researchers. Many knowledge representation techniques (Chakraborty (2008); Karabatak and Ince (2009); Sittig et al. 2010; Tanwar et al. 2010a,b; Arsene et al. 2015) are available in the literature. Among them, production rules are extensively used in medical domain (Kong et al. 2008) and resemble as a good starting point for real-life understanding of the expert systems (Gamberger et al. 2008). Since 1970s, rule-based knowledge representation is being used in many medical expert systems such as MYCIN (Shortliffe et al. 1975; Shortliffe (1976); Shortliffe (2012)), PUFF (Aikins et al.1983), UMLS-based CDSS (Achour et al. 2001), CMDS (Huang and Chen (2007)), and so on (Liao (2005); Kuo et al. 2011). Sometimes, production rules have been associated with some probabilistic measures to deal with uncertainties in medical knowledge. Resolution of clinical uncertainties using fuzzy logic (Klir and Yuan (1995); Yen and Langari (1999)) and Bayesian Networks (Friedman et al. 1997; Friedman et al. 1999) is quite popular. Rough Set has also proved its effectiveness in handling inconsistencies in medical knowledge (Tsumoto (1998); Mitra et al. 2006; Ilczuk et al. 2005; Hassanien et al. 2009; Tsumoto (2004)). In this section, we first discuss about knowledge representation techniques in some medical expert systems, and then we discuss some medical applications where rough set has been used. Finally we show how lattice-based structure can be used for knowledge representation.

The production rules have been used for knowledge representation in MYCIN (Shortliffe et al. 1975; Shortliffe (1976); Shortliffe (2012)). Each of the rules which has a premise (Boolean combination of predicate functions) and an action, is associated with a metric called 'Certainty Factor' which indicates the strength of a rule. Frame-based knowledge representation has been used by (Langlotz et al. 1988) to describe particular objects or concepts. Using the frames, a decision tree has been constructed for reasoning. Many algebraic relationships exist between the tree - nodes in the decision tree and different medical situations. The qualitative knowledge used for analysis of the expressions is represented using production rules.

The developers of ONCOCIN (Shortliffe et al. 1984) used three kinds of data structures for knowledge representation: control blocks, frames, and rules. While the control blocks were used for encoding procedural knowledge to care for cancer patients, frames were used for storing static protocol knowledge. On the other hand, rules had been used for data-driven or goal-driven inference. ONCOCIN defined a list of contexts for the production rules to be applicable. The modular approach for knowledge representation achieved by developing three different data structures made ONCOCIN more efficient during consultations.

(Weiss et al. 1978) have developed a causal-association network (CASNET) model for capturing the pathophysiological mechanisms of Glaucoma. The knowledge has been represented using three types of data elements corresponding to observations of a patient, states of the causal net, as well as the diagnostic, prognostic, and therapeutic categories. The network produces different types of causal rules with some causal weightages called confidence factors. The rules are logically ordered to produce diagnostic, prognostic and explanatory conclusions. While CASNET model simply propagates the weights among the causal concepts, another approach called ABEL (Patil et al. 1981) is used to symbolically manipulate the weights through various operations. ABEL uses a hierarchical representation of medical knowledge, with the lowest level describing pathophysiological knowledge about diseases, which is aggregated into higher level concepts and relations to ultimately produce syndromic knowledge. This kind of knowledge representation can capture any domain knowledge in different levels of detail.

Case-based reasoning has been used for knowledge representation in many applications (Holt et al. 2005; Bichindaritz et al. 1998; Bichindaritz and Marling (2006); Schmidt et al. 2001) in order to extract knowledge easily from expert physicians in the form of cases. There is also medical application of ontology-driven knowledge representation (Jovic et al. 2007), which holds relationships between concepts such as treatment, clinical procedures, patient data etc. Also production rules can be generated from ontology. (Gamberger et al. 2008) have developed a knowledge base whose descriptive part is designed with ontology and the procedural part has been designed using production rules.

Rough set has been used for knowledge discovery (Polkowski (2013); Huang and Tseng (2004)) in many medical applications. (Own and Abraham (2012)) has proposed a rough set based framework for early intervention and prevention of neurological dysfunction and kernicterus that lead to Egyptian neonatal jaundice. The authors have designed a weighted information table, extracted relevant attributes from the information table, formed a reduced set of weighted attributes, and finally achieved a set of diagnosis rules applicable for neonatal jaundice. A rule-based rough-set decision support system has been designed for ECG classification by (Mitra et al., 2006). Here rough set theory has been used for handling different kinds of inconsistencies occurred due to transcription errors in ECG signals, subjective calculation of attribute values, lack of information etc. This rough-set decision support system is expected to produce low-cost, more robust, and human-like decisions.

Knowledge can also be represented using concept lattice (Zhang et al. 2005; Belohlavek (2008); Aswani Kumar and Srinivas (2010)), which follows lattice structure. There exist some notable works that combine concept lattice and rough set theory (Yang and Xu (2009); Wei and Qi (2010); Yao (2004); Tripathy et al. 2013; Saquer and Deogun (2001)). Some theories have been proposed for combination of rough set theory and lattice theory (Järvinen (2007); Praba and Mohan (2013); Estaji et al. 2012; Xiao et al., 2014), though the application for the combined theory in medical domain is very limited.

Lattice theory, associated with rough set, can be applied in medical domain in many different ways. In this paper, we have applied rough-lattice theory for generation of exhaustive but optimal set of decision rules. The working principle of concept lattice may not always follow the guidelines of a medical domain, as combinations of different clinical symptoms may indicate a new set of diseases. Pure lattice-based model built on symptoms would be advantageous in this regard. Also, to cope with exponential growth of medical knowledge, an incremental design of knowledge base should be invoked. Generation of decision rules from dynamic databases using incremental rough set approach in less time compared to the other approaches (Blaszczynski and Słowiński (2003); Asharaf et al. 2006; Guo et al. 2005) has been proposed by (Fan et al. 2009).

Our approach uses a lattice of raw knowledge as an information system for rough set, instead of using conventional tabular information system. Though a number of literature exists on rough-lattice theory, its application in medical knowledge representation is rarely found. Compared to the existing approaches, the advantage of using lattice for medical knowledge representation is three-fold: first, the lattice, being a directed graph, is a good way of knowledge representation that shows clearly how every piece of knowledge is related to each other; second, lattice captures an exhaustive set of knowledge in a well-organized and optimized way ensuring completeness and zero-redundancy; third, operations like insertion, deletion, and modification on lattice structure are easy. As knowledge is acquired from different knowledge sources in incremental fashion, overcoming the problem of redundancy is a great achievement as the knowledge base can be stored in optimal space. Therefore the computational time for processing of the stored knowledge would be less. When rough set uses this lattice-based information system, the generation of optimal diagnostic rules becomes much easier. Overall, the rough set based lattice structure for knowledge representation appears as a computationally efficient technique with respect to both time and space. With the introduction of new reliability metric, the proposed technique is expected to offer more reliability in diagnostic decision making compared to the already-available medical expert systems.

## 3. Background study

### 3.1 Preliminaries of lattice theory

A Relation $R$ over any set $X$ is a subset of $X \times X$. A relation $R$ is reflexive if for each $x \in X$, $(x, x) \in R$. $R$ is irreflexive if $(x, x) \notin R$. A relation $R$ is symmetric if for all $x, y \in X$, $(x, y) \in R$ implies $(y, x) \in R$. $R$ is antisymmetric if for all $x, y \in X$, $(x, y) \in R$ and $(y, x) \in R$ implies $x = y$. A relation $R$ is asymmetric if for all $x, y \in X$, $(x, y) \in R$ implies $(y, x) \notin R$. A relation $R$ is transitive if for all $x, y, z \in X$, $(x, y) \in R$ and $(y, z) \in R$ implies $(x, z) \in R$. A relation $R$ is an equivalence relation if it is reflexive, symmetric and transitive. Suppose $(x, y) \in R$ and $R$ is an equivalence relation. We denote this phenomenon by $x \equiv^R y$. Furthermore, for each $x \in X$, equivalence class of $x$, denoted by $[x]^R$, is defined as the set of all $y \in X$ such that $y \equiv^R x$. The set of all such equivalence classes forms a partition of $X$.

A relation $R$ is a partial order if it is reflexive, antisymmetric and transitive. When $R$ is a partial order, we use $x \leq y$ to denote $(x, y) \in R$. The set $X$ together with the partial order is denoted by $(X, \leq)$. We use $P = (X, \leq)$ to denote a partially ordered set or 'poset'. Suppose $Y \subseteq X$. We now define two operators on subsets of the set $X$: 'meet' or 'infimum' and 'join' or 'supremum'. For any $m \in X$, we say that $m$ = meet $Y$ iff $\forall y \in Y$: $m \leq y$, and $\forall m' \in X$: $(\forall y \in Y$: $m' \leq y) \Rightarrow m' \leq m$. The first condition says that $m$ is a lower bound of the set $Y$. The second condition says that if $m'$ is another lower bound of $Y$, then it is less than $m$. For this reason, $m$ is also called the greatest lower bound ($glb$) of the set $Y$. It is easy to check that the infimum of $Y$ is unique whenever it exists. However, $m$ is not required to be an element of $Y$. The definition of 'join' is similar. For any $s \in X$, we say that $s$ = join $Y$ iff $\forall y \in Y$: $y \leq s$, and $\forall s' \in X$: ($\forall y \in Y$: $y \leq s'$) $\Rightarrow s \leq s'$. Here, $s$ is also called the least upper bound (*lub*) of the set $Y$. We denote the *glb* of $\{a, b\}$ by $(a \wedge b)$, and *lub* of $\{a, b\}$ by $(a \vee b)$. The greatest lower bound or the least upper bound may not always exist. A poset $(X, \leq)$ would be called a lattice iff $\forall x, y \in X$, the *glb* $(x \wedge y)$ and the *lub* $(a \vee b)$ always exist.

Finite posets are often depicted graphically using Hasse diagram. To define a Hasse diagram, we first define a relation cover as follows. For any two elements $(x, y) \in X$, $y$ covers $x$ if there is no element $z$ ($\forall z \in X$) with $x \leq z \leq y$. In this case, $y$ is an upper cover of $x$ and $x$ is a lower cover of $y$. A Hasse diagram of a poset is a graph with the property that there is an edge from $x$ to $y$ iff $y$ covers $x$. Furthermore, while drawing the graph on a Euclidean plane, $x$ is drawn lower than $y$ when $y$ covers $x$. This allows us to suppress the directional arrows in the edges.

### 3.2 Basics of rough set theory

Rough set theory, proposed by Pawlak in 1982, is a mathematical approach to deal with inconsistency, imprecision, and incompleteness in knowledge. The rough set theory assumes that every object in the universe is associated with some information or knowledge. As an example, we consider three diseases CFJ (Cervical Facet Joint pain) (Manchikanti et al. 2004), discogenic pain (Saal and Saal (2000)), and myofacial pain syndrome (MPS) (Malanga and Colon (2010)) as three objects from the LBP domain. Two symptoms namely "LBP without radiculopathy" and "no LBP at rest" are common in all the three diseases. So, we can say that the three diseases are characterized by these two symptoms. Therefore, these two symptoms are basically the information associated with the three diseases. In this case, as the three diseases are characterized by the same information, these three diseases are called indiscernible in view of the available information (i.e. the two symptoms under consideration) about them.

The basis of rough set theory is an information system. Formally, an information system may be expressed as a pair $S = (U, A)$, where $U$ and $A$ are finite nonempty sets called the 'universe' and the set of 'attributes', respectively. Every attribute $a \in A$ is associated with a set $V_a$, of its values, called the domain of $a$. Any subset $B$ of $A$ determines a binary relation $I(B)$ on $U$, which will be called an indiscernibility relation. The indiscernibility relation is an equivalence relation. Equivalence classes of the relation $I(B)$ are referred to as B-elementary sets or B-granules. Any finite union of elementary sets is called a *definable set*. If a set of attributes and its superset define the same indiscernibility relation (i.e. if elementary sets of both relations are identical), then any attribute that belongs to the superset and not to the set is redundant. A set of attributes with no redundant attribute is called minimal (or independent). The set $P$ of attributes is the *reduct* (or *covering*) of another set $Q$ of attributes if $P$ is minimal and the indiscernibility relations defined by $P$ and $Q$ are the same.

The attributes of an information system may further be classified as 'condition' and 'decision' attributes. So, the previously designed information system is modified as $S = (U, C, D)$, where $C$ and $D$ are disjoint sets of condition and decision attributes respectively. Here, $C \cup D = A$. This kind of information system is called a *decision table*. If we take an element $x \in U$, we can induce a decision rule as follows: $[c_1(x), c_2(x), \ldots c_n(x)] \rightarrow [d_1(x), d_2(x), \ldots d_m(x)]$ or $C \rightarrow_x D$ ,where $C = \{c_1, c_2, ..., c_n\}$ and $D = \{d_1, d_2, ..., d_m\}$. Associated with each decision rule $C \rightarrow_x D$, there are four quality metrics: *support*, *strength*, *certainty factor*, and *coverage factor*. Decision rules are often represented in a form of "If *condition* Then *decision*". Thus, any decision table can be transformed in a set of "If *condition* Then *decision*" rules, from which a set of optimal rules is generated.

Now, suppose $X \subseteq U$, and $B \subseteq A$. We need to characterize set $X$ in terms of attributes from $B$. If we cannot be able to induce a precise/crisp description of elements of set $X$ using $B$, we have to find out the lower and upper approximations of set $X$. So, generation of lower and upper approximations is required to deal with inconsistency in knowledge. The $B$ - Lower Approximation of $X$ (denoted by $B_*(X)$) and $B$-Upper Approximation of $X$ (denoted by $B^*(X)$) are formally defined as follows: $B_*(X) = \cup_{x \in U} \{B(x): B(x) \subseteq X\}$ and $B^*(X) = \cup_{x \in U} \{B(x): B(x) \cap X = \emptyset\}$.The set containing elements from the upper approximation of a concept that are not members of the lower approximation of the concept, is called a boundary region. The set $BN_B(X)$ [= $(B^*(X) - B_*(X))$] is called the $B$-Boundary Region of $X$ and thus consists of those objects that we cannot decisively classify into $X$ on the basis of knowledge in $B$. If $BN_B(X) = \emptyset$, then $X$ is Crisp (exact) w.r.t. $B$. If $BN_B(X) \neq \emptyset$, then $X$ is Rough (inexact) w.r.t. $B$. A Rough Set may be defined as a set having non-empty boundary regions.

## 4. Proposed knowledge representation technique

We conceptualize the knowledge as a collection of rules, with each rule containing a condition part (*C*) and a decision part (*D*). *C* consists of $n$ (> 0) facts and *D* consists of $m$ (> 0) decisions. We denote an attribute-value pair as a fact, and a disease-value pair as a decision. In this work, condition parts hold sets of symptoms and decision parts include sets of diseases. Mathematically we can express a rule *R* as follows:

$$R: C \rightarrow D, \text{ where } C = \prod_{k=1}^{n} f_k \quad \text{and } D = \sum_{l=1}^{m} D_l, \text{ where } f_k \text{ is the } k^{th} \text{ fact, } D_l \text{ is the } l^{th} \text{ decision.}$$

Here $f_k = (Attribute_k,\ Value)$ and $D_l = (d_l, V_d)$. $V_d$ is a ternary variable which may accept 3 types of values: *true* (1), *false* (0), or *inconclusive* (2). When $V_d$ =1 for $d_l$, we say that the disease surely occurs for a particular set of symptoms. If $V_d$ = 0 for $d_l$, we can conclude that the disease never occurs for a particular set of symptoms. Again, if $V_d$ = 2 for $d_l$, we would say that for a particular set of symptoms, $d_l$ may or may occur. A rule which contains a single fact is called an atomic rule. A rule which contains more than one fact is called a composite rule. We design the knowledge base (KB) as a set of such kinds of rules.

We now build a lattice structure for capturing the unrefined rules gathered directly from the knowledge sources. The lattice of such rules would act as the information system for rough set. The rough set will then remove the inconsistent knowledge hidden in the lattice and would produce a set of optimal decision rules. Figure 1 depicts the proposed technique.

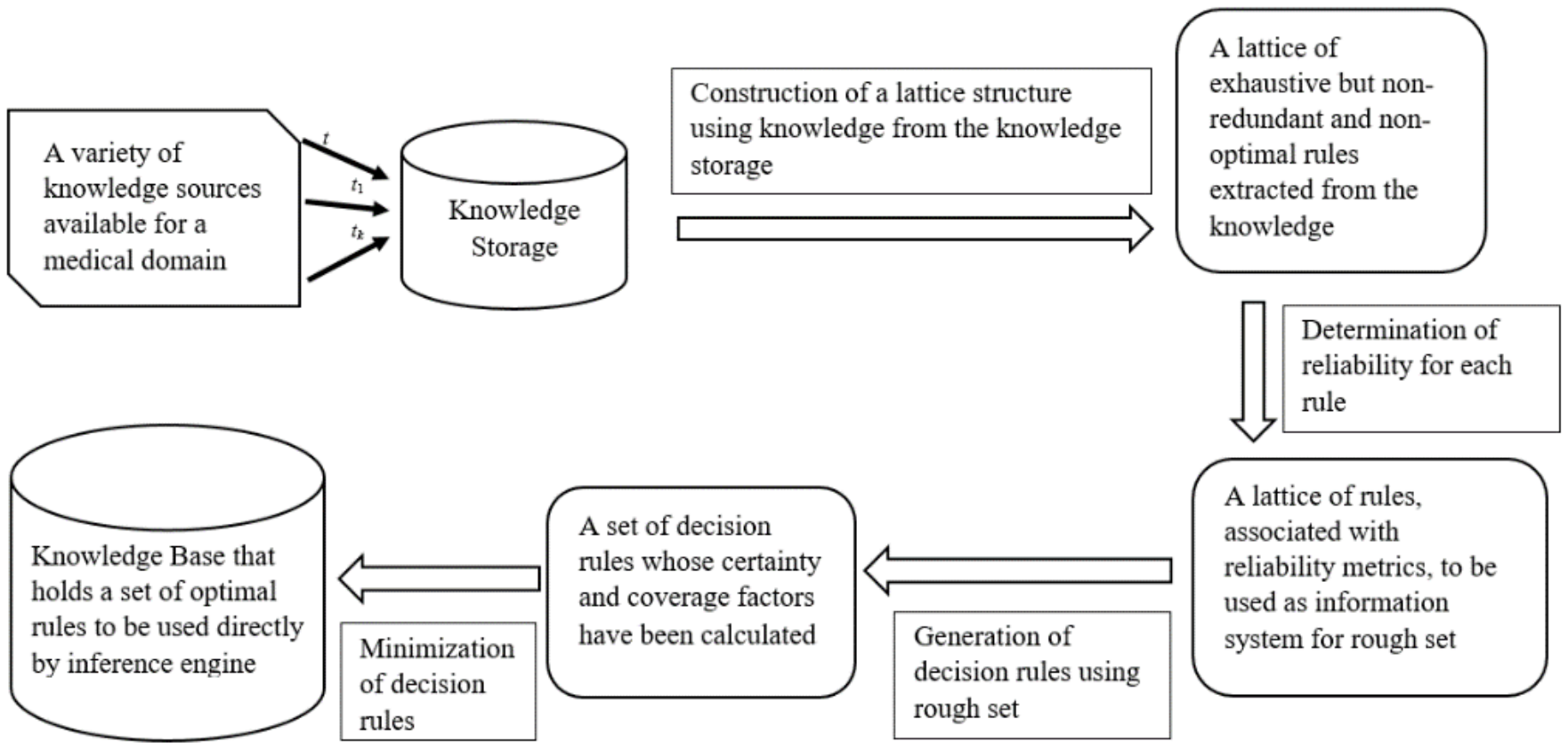

**Figure 1**. Rough set based lattice structure for medical knowledge representation

### 4.1 Lattice structure for knowledge representation

Initially KB is empty i.e. KB = $\phi$. Every time we encounter a new rule, we add it to the existing KB using 'set union' operation. We first insert the atomic rules in KB. Suppose we have $n$ (> 0) atomic rules. So, the knowledge base is modified as KB = KB $\cup$ $\{R_i \mid R_i$ is an atomic rule with fact $f_i$ where $1 \leq i \leq n\}$. Then we insert composite rules in KB. For this, we combine the facts of the atomic rules in the following way: first we construct the total number of possible groups that can be formed by taking two different facts at a time out of $n$ distinct facts. Obviously the total number of such groups is ${}^nC_2$. So, from $n$ atomic rules, we can generate at most ${}^nC_2$ distinct composite rules whose condition parts would contain exactly two facts. These rules are included in KB. Now we form groups of facts taking three different facts at a time from the set of $n$ facts and generate composite rules with three facts accordingly. The new knowledge is inserted into the KB. This process continues until we reach to a state where only one group is formed taking $n$ distinct facts at a time. So, at this stage, only one composite rule with $n$ facts would be generated. After the construction process is complete, the KB would hold $(2^n - 1)$ rules. If we consider only the sets of facts in KB, we obtain a set *S* which includes $2^n$ combinations of facts including $\phi$. So, the set $S = \{\phi, \{f_1\}, \{f_2\}, \{f_3\}, \ldots, \{f_n\}, \{f_1, f_2\}, \{f_1, f_3\}, .., \{f_{n-1}, f_n\}, \{f_1, f_2, f_3\}, \ldots, \{f_{n-2}, f_{n-1}, f_n\}, \ldots \{f_1, f_2, \ldots, f_{n-1}, f_n\}\}$.

Now we consider a relation *R* as subset equality i.e. $\subseteq$. Suppose $A, B, C \subseteq S$. We say that $[S; \subseteq]$ is a poset as

i) $\forall A \in S : (A \subseteq A)$ (holds reflexivity relationship)
ii) $\forall (A, B) \in S : A \subseteq B$ and $B \subseteq A \Rightarrow (A = B)$ (holds antisymmetry relationship)
iii) $\forall (A, B, C) \in S : A \subseteq B, B \subseteq C \Rightarrow (A \subseteq C)$ (holds transitivity relationship)

Now, we take two elements $x$ ($\in S$) and $y$ ($\in S$) and construct $\{x, y\}$. This set would always have a *glb* as $(x \cap y)$ ($\in S$) and *lub* as $(x \cup y)$ ($\in S$). This is true for any $x$ ($\in S$) and $y$ ($\in S$). So, $S$ is a lattice of order $n$.

If we draw a Hasse diagram for lattice $S$, it is easy to visualize that $S$ follows a bottom-up design starting from the set $\phi$ at level 0 (bottom-most level). $\phi$ would be related to singleton sets $\{f_1\}, \{f_2\}, \{f_3\},\ldots,\{f_n\}$ at level 1. Sets at level 1 would be related to sets of two elements $\{f_1, f_2\}, \{f_1, f_3\},..,\{f_{n-1}, f_n\}$ in such a manner that if $\{f_i\}$ and $\{f_j\}$ are two distinct sets at level 1, then they would be related to a set $\{f_i, f_j\}$ at level 2. The strategy to proceed upwards is as follows: $\forall (L_1, L_2) \in S$ where $|L_1| = |L_2| = k$ ($> 0$), $L_1$ and $L_2$ would be related to $(L_1 \cup L_2)$ ($\in S$) and $|L_1 \cup L_2| = k + 1$. Here $L_1$ and $L_2$ are at $k^{th}$ level and $(L_1 \cup L_2)$ is at $(k + 1)^{th}$ level. We stop when we obtain the set $\{f_1, f_2, \ldots, f_n\}$ at level $n$. So, a lattice of order $n$ has $(n + 1)$ levels.

As $S$ is obtained from KB, KB follows a lattice structure on the condition parts of the rules. Suppose in KB, $R_i$: $C_i \rightarrow D_i$ is an atomic rule, where $C_i$ contains only fact $f_i$, and $R_j$: $C_j \rightarrow D_j$ is another atomic rule, where $C_j$ contains fact $f_j$ ($j \neq i$). As the lattice $S$ says, $\{f_i\}$ and $\{f_j\}$ would be related to $\{f_i, f_j\}$ leading to induction of a new rule

$R_{ij}$: $(C_i \cup C_j) \rightarrow D$ $[\subseteq (D_i \cup D_j)]$, if no new set of diseases exists for combination of $C_i$ and $C_j$
$\rightarrow (D \cup D_{ij})$, where $D_{ij}$ is the new set of decisions for combination of $C_i$ and $C_j$

While symptoms $C_i$ and $C_j$ imply a set of decisions $D_i$ and $D_j$, symptoms $C_i$ combined with $C_j$ may indicate a new set of diseases $D_{ij}$ ($\neq (D_i \cup D_j)$). Occurrence of a new set of diseases for a combination of symptoms is very common in medical field. Non-consideration of any decision in the induced rule at any moment, without any strong evidence, could make a fatal consequence for patient. For example, if a female patient comes with LBP and fever, then, besides considering the individual causes, there is a possibility that the patient may suffer from some bacterial infections, tuberculosis or cancer (Medappil and Adiga (2011)). It is also frequent in medical domain that, from a single symptom no diagnostic decision can be made. This kind of phenomenon is captured in our KB using atomic rules with empty decision parts.

#### 4.1.1 Quantifying factor for a rule

A disease may be caused by more than one symptom. But influence of each symptom behind occurrence of a diseases may not be same. So, for a disease to occur, symptoms may have different contributory weightages. For example, if an LBP patient has localized muscle pain with observable trigger points, then the patient may be suffering from MPS. In this case, the chance of the patient having MPS is very high. So, this symptom, being independent of other symptoms, carries a huge weightage for leading to the disease MPS. Assume that the patient also shows jump sign on palpation at a trigger point. Jump sign on palpation is an obvious indicator of MPS but the reverse is not always true (Malanga and Colon (2010)). So, this symptom carries lower weightage. Consider a rule $R$ whose condition part includes two facts $f_1$ and $f_2$, where $f_1$ = ("Localized muscle pain with observable trigger points", "Yes") and $f_2$ = ("Jump sign on palpation at trigger point", "Yes"). The decision part includes the disease MPS with $V_d = 1$. The different weightages of symptoms for a particular decision is termed as 'Conditional Weightages' ($C_W$). In this rule, conditional weightage for $f_1$ is high and that of $f_2$ is low. It may also be that all the facts in a rule carry the same conditional weightage.

We now try to define the conditional weightage of a fact mathematically. Suppose there is a rule which includes $n(> 0)$ facts with priorities $p(\leq n)$ for a particular disease $d$. A fact $f_i$ has priority $p_i$. The priority of each fact is obtained from different knowledge sources. We define the conditional weightage $C_w^i$ associated with each fact $f_i$ ($1 \leq i \leq n$) for disease $d$ as $C_w^i = p_i/n$.

Now we define a quantifying factor for a rule keeping in mind the uncertainties in a medical domain. A rule can be formally represented using 8-tuple $<C, C_w, d, T_v^1, T_v^2, T_v^3, V_d, CF>$ , where $C = \{ f_i \mid f_i$ is a fact and $1 \leq i \leq n\}$; $C_w = \{C_w^i \mid C_w^i$ is the conditional weightage of $f_i\}$; $d$ is the disease for the pair $<C, C_w>$; $T_v^1$ denotes a value to measure the acceptability of the knowledge sources that give evidence about the certain presence of disease *d* for <*C*, $C_w$>; $T_v^2$ denotes a value to measure the acceptability of knowledge sources that provide evidence in support of certain absence of disease *d* for the same pair; $T_v^3$ denotes a value to measure the acceptability of knowledge sources which tell us that the disease *d* may or may not be present in the current context. Each of the values would be between 0 and 1; $V_d$ is the truth value for *d* which is decided based on $T_v^1$, $T_v^2$, $T_v^3$; *CF* (Credibility Factor) gives some measure of the rule for being believable or trustworthy. One important thing to mention here is that if $R$: $C \rightarrow D$ is a rule where $D$ = {<$d_k$, $V_d$> | <$d_k$, $V_d$> is a decision and $1 \le k \le K$}, we break down *R* into *K* sub-rules of the form $R_k$: $C \rightarrow$ <$d_k$, $V_d$>. Each sub-rule also follows the 8-tuple representation.

Acceptability of a knowledge source is measured based on the nature of the source. According to opinions of different experts from the domain of LBP, the various sources of knowledge have been assigned weightages as per their acceptability levels; e.g. national guidelines have been graded as level 1(highest priority), randomized controlled trials (RCT) (Stolberg et al. 2004) is graded as level 2, existing literature (journals/ articles/ books etc.) belong to level 3 category, patient case studies belong to level 4 category, and recommendations of expert physicians belong to level 5 category (lowest priority). As this kind of gradation is not standardized, we say that, in general, knowledge resources can be classified into *q* different categories, where level 1 means highest priority and level *q* is considered as lowest priority.

The generalized formula for calculating $T_v^m$ (where $1 \le m \le 3$) is as follows: we consider that there are $K_m$ knowledge sources corresponding to $T_v^m$. We assume that all the knowledge sources of set $K_m$ are not of same category, with a set of $|K_{mj}|$ ($K_{mj} \subseteq K_m$) sources belonging to level *j* category, where $1 \le j \le q$. The consideration is that $K_{m1} \cap K_{m2} \cap \ldots \cap K_{mq} = \phi$ and $K_{m1} \cup K_{m2} \cup \ldots \cup K_{mq} = K_m$. The generalized forms of $T_v^1$, $T_v^2$, and $T_v^3$ are given in equation (i).

$$T_v^1 = \frac{x}{W}, T_v^2 = \frac{y}{W}, T_v^3 = \frac{z}{W} \qquad \text{………….. (i)}$$

where $x = (\, q|K_{11}| + (q-1)|K_{12}| + (q-3)|K_{13}| + \cdots + 2|K_{1(q-1)}| + |K_{1q}|)$

$y = (\, q|K_{21}| + (q-1)|K_{22}| + (q-3)|K_{23}| + \cdots + 2|K_{2(q-1)}| + |K_{2q}|)$

$z = (\, q|K_{31}| + (q-1)|K_{32}| + (q-3)|K_{33}| + \cdots + 2|K_{3(q-1)}| + |K_{3q}|)$

$W = (x + y + z)$

The implicit assumption is that $W \neq 0$. The complemented forms of the parameters are as follows:
$\sim T_v^1 = 1 - T_v^1$, which reflects how much unacceptability is there to claim certain presence of disease *d* for *C*;
$\sim T_v^2 = 1 - T_v^2$, which reflects how much unacceptability is there to claim certain absence of disease *d* for *C*;
$\sim T_v^3 = 1 - T_v^3$, which reflects how much unacceptability is there to say that disease *d* may or may not be present.

As we have just seen that a decision is taken based on conditions with different conditional weightages, measuring whether the rule is trustworthy requires careful investigation about different parameters like $T_v^m$, $K_{m1}$, …, $K_{mq}$. We now construct a (3×*q*) matrix *M* where, the first row is for $K_1$, second row is for $K_2$, and third row is for $K_3$. The columns 1 to *q* represent different categories of knowledge sources. Our assumption is that indices of *M* start from location (1,1). $M(x,y) = 1$ if for row $x$ ($1 \le x \le 3$), there are knowledge sources under category $y$ ($1 \le y \le q$), else $M(x,y) = 0$. Using *M*, we determine $V_d$ of *d* and also calculate the corresponding *CF*.

If 1 is present at one or more than one column of only one row *x* of *M*, then $V_d = 1$ and $CF = T_v^1$ if $x = 1$; $V_d = 0$ and $CF = T_v^2$ if $x = 2$; $V_d = 2$ and $CF = T_v^3$ if $x = 3$. Now, consider that 1 is present in multiple rows of *M*, then we proceed column-wise from left to right checking for first $y$ ($\ge 1$) when $M(x, y) = 1$. So, obviously $M(x_1, y_1) = 0$ where $1 \le y_1 < y$ and $1 \le x_1 \le 3$. We shall now discuss about two cases: i) $M(x, y) = 1$ for a single row *x* of *M* and for each $x'$ representing the rest two rows, $M(x', y) = 0$; ii) $M(x, y) = 1$ for each *x* from more than one row.

The first case is very straight-forward. In this case, $V_d = 1$ and $CF = T_v^1$ if $x = 1$; $V_d = 0$ and $CF = T_v^2$ if $x = 2$; $V_d = 2$ and $CF = T_v^3$ if $x = 3$. The second case requires considering situations when two rows have value 1 under column *y* and also when three rows of *M* have value 1 at column *y*. So, there are basically four such combinations as given below:

a) **Case:** $M(1,y) = 1$, $M(2,y) = 1$, and $M(3,y) = 0$

If $T_v^1 > T_v^2$ , then $V_d = 1$ and $CF = T_v^1$. Else if $T_v^1 < T_v^2$ , then $V_d = 0$ and $CF = T_v^2$. But if $T_v^1 = T_v^2$, then starting from column $(y+1)$, we would find the first $y'$ ($y < y' \leq q$) where either $M(1, y') = 1$ or $M(2, y') = 1$, but not both. If $M(1, y') =1$, then $V_d = 1$ and $CF = T_v^1$; else $V_d = 0$ and $CF = T_v^2$. If no such $y'$ is found in $M$, $V_d = 2$ and $CF = T_v^1$.

b) **Case:** $M(1,y) = 1$, $M(2,y) = 0$, and $M(3,y) = 1$
If $T_v^1 > T_v^3$, then $V_d = 1$ and $CF = T_v^3$. Else if $T_v^1 < T_v^3$ , then $V_d = 2$ and $CF = T_v^3$. But if $T_v^1 = T_v^3$, then starting from column $(y+1)$, we find the first $y'$ ($y < y' \leq q$) where either $M(1, y') = 1$ or $M(3, y') = 1$, but not both. If $M(1, y') =1$, then $V_d = 1$ and $CF = T_v^1$; else $V_d = 2$ and $CF = T_v^3$. If no such $y'$ is found in $M$, $V_d = 2$ and $CF = T_v^3$.

c) **Case:** $M(1,y) = 0$, $M(2,y) = 1$ and $M(3,y) = 1$
If $T_v^2 > T_v^3$, then $V_d = 0$ and $CF = T_v^2$. Else if $T_v^2 < T_v^3$ , then $V_d = 2$ and $CF = T_v^3$. But if $T_v^2 = T_v^3$, then starting from column $(y+1)$, we find the first $y'$ ($y < y' \leq q$) where either $M(2, y') = 1$ or $M(3, y') = 1$, but not both. If $M(2, y') = 1$, then $V_d = 0$ and $CF = T_v^2$; else $V_d = 2$ and $CF = T_v^3$. If no such $y'$ is found in $M$, $V_d = 2$ and $CF = T_v^3$.

d) **Case:** $M(1,y) =1$, $M(2,y) = 1$, and $M(3,y) = 1$
If $T_v^1 > T_v^2$ and $T_v^1 > T_v^3$ then $V_d = 1$ and $CF = T_v^1$. Else if $T_v^1 < T_v^2$ and $T_v^2 > T_v^3$ , then $V_d = 0$ and $CF = T_v^2$. If $T_v^3 > T_v^1$ and $T_v^3 > T_v^2$ , then $V_d = 2$ and $CF = T_v^3$. For other kinds of relationships among $T_v^1, T_v^2, T_v^3$ except any equality relation, we find out the maximum of all three and determine $V_d$ and $CF$ accordingly. Now, consider that equality relation exists between $T_v^1$ and $T_v^2$ and $T_v^1 < T_v^3$. In this case $V_d = 2$ and $CF = T_v^3$. Similar strategy would be followed if $T_v^1 = T_v^3$ and $T_v^1 < T_v^2$, or $T_v^2 = T_v^3$ and $T_v^2 < T_v^1$. For the reverse case, when $T_v^1 = T_v^2$ and $T_v^1 > T_v^3$, we say $V_d = 2$ and $CF = T_v^1$. For the rest cases, $V_d = 2$ and $CF = T_v^3$.
If $T_v^1 = T_v^2 = T_v^3$, then starting from column $(y+1)$, we find the first $y'$ ($y < y' \leq q$) where either $M(1, y') = 1$ or $M(2, y') = 1$ or $M(3, y') = 1$, but not all, provided that no two rows have 1s under column $y''$ ($y < y'' < y'$). If $M(1, y') =1$, then $V_d = 1$ and $CF = T_v^1$; if $M(2, y') =1$, then $V_d = 0$ and $CF = T_v^2$; else $V_d = 2$ and $CF = T_v^3$. If no such $y''$ and $y'$ are found in $M$, $V_d = 2$ and $CF = T_v^3$. But if we have $M(1, y'') = 1$ and $M(2, y'') = 1$, we then search for the next column $y^+$ ($y'' < y^+ \leq q$), where $M(x^+, y^+) = 1$ with $x^+$ being the first or second row. If $M(1, y^+) =1$, then $V_d = 1$ and $CF = T_v^1$. If $M(2, y^+) =1$, then $V_d = 0$ and $CF = T_v^2$. If no such $y^+$ is found in $M$, $V_d = 2$ and $CF = T_v^1$.We also act in similar way when $M(1, y'') =1$ and $M(3, y'') =1$. If $M(1, y^+) =1$, then $V_d = 1$ and $CF = T_v^1$. If $M(3, y^+) =1$, then $V_d = 2$ and $CF = T_v^3$. If no such $y^+$ is found in $M$, $V_d = 2$ and $CF = T_v^3$. Likewise, for $M(2, y'') =1$ and $M(3, y'') =1$, if $M(2, y^+) =1$, then $V_d = 0$ and $CF = T_v^2$; else $V_d = 2$ and $CF = T_v^3$. If no such $y^+$ is found in $M$, $V_d = 2$ and $CF = T_v^3$.

Now let us consider two atomic rules $R_i$: $C_i \rightarrow D_i$ and $R_j$: $C_j \rightarrow D_j$, where $i \neq j$. Suppose $D_i = (d, V_d^i)$ and $D_j = (d, V_d^j)$. From these two rules, a new rule $R_{ij}$: $C_{ij} \rightarrow D_{ij}$ is induced, where the conditional weightage for fact $f_i$ is $C_w^i$ and that of fact $f_j$ is $C_w^j$. The conditional weightages are supplied by knowledge sources. If there is no knowledge source available for $R_{ij}$, then we would assume that the conditional weightages are same for each fact. As per our consideration, $D_{ij}$ would also include $d$ as the disease. If knowledge sources are available for $R_{ij}$, we consider that $K_{ij}^1$ is the set of knowledge sources to denote certain presence of disease $d$ for $(C_i \cup C_j)$, $K_{ij}^2$ is the set of knowledge sources to denote certain absence of disease $d$, and $K_{ij}^3$ is the set of knowledge sources which indicate that disease $d$ may or may not be present for $(C_i \cup C_j)$. We now calculate $T_v^m$ $[1 \leq m \leq 3]$ for $R_{ij}$, denoted by $[T_v^m]_{ij}^*$, as per equation (i). Note that, $[T_v^m]_{ij}^*$ is obtained from external knowledge sources for $R_{ij}$ and it is not the actual value of $T_v^m$. If no such knowledge source is available, $[T_v^m]_{ij}^*$ would be 0. The actual value of $T_v^m$ for $R_{ij}$, denoted by $[T_v^m]_{ij}$, is obtained as $([T_v^m]_i + [T_v^m]_j + [T_v^m]_{ij}^*)/3$, where $[T_v^m]_i$ and $[T_v^m]_j$ represent the values of $T_v^m$ for $R_i$ and $R_j$ respectively. If, $[T_v^m]_{ij}^* = 0$, then $[T_v^m]_{ij} = ([T_v^m]_i + [T_v^m]_j)/2$.

Now we determine the $V_d$ of $d$ and $CF$ of $R_{ij}$ based on the available knowledge of $R_i$ and $R_j$ in the lattice-based KB. The values obtained from the implicit lattice knowledge are termed as $[V_d^*]_{ij}$ and $CF^*(R_{ij})$. If no knowledge sources are available for $R_{ij}$, the final $V_d$ of $d$ in $R_{ij}$ is $[V_d]_{ij} = [V_d^*]_{ij}$ and the final $CF$ of $R_{ij}$ $CF(R_{ij}) = CF^*(R_{ij})$.

**Case 1: Two constituent rules having the same $V_d$ for $d$**

Suppose two rules $R_i$ and $R_j$ hold the same truth value for disease $d$. We say that the newly induced rule $R_{ij}$ would also have the same truth value. The motivation behind this consideration is that, if father and mother carry the same genetic characteristic, the child also would inherit the same genetic characteristic. We now calculate the credibility factor for $R_{ij}$ against the decision $D_{ij}$ as shown in equation (ii). Here, we use a threshold $\alpha$ $(0 \le \alpha \le 1)$ if we want to restrict use of the credibility factor of a rule in some further calculations. The value of $\alpha$ is application-specific.

$$CF^*(R_{ij}) = \begin{cases} (CF(R_i)C_w^i + CF(R_j)C_w^j), & CF(R_i)C_w^i > \alpha \text{ and } CF(R_j))C_w^j > \alpha \\ CF(R_i)C_w^i, & CF(R_j)C_w^j \le \alpha \text{ and } CF(R_i)C_w^i > \alpha \\ CF(R_j)C_w^j, & CF(R_i)C_w^i \le \alpha \text{ and } CF(R_j)C_w^j > \alpha \\ 0, & CF(R_i)C_w^i \le \alpha \text{ and } CF(R_j)C_w^j \le \alpha \end{cases} \quad \text{........... (ii)}$$

**Case 2: Two constituent rules having different $V_d$ for $d$**

Two constituent rules $R_i$ and $R_j$ may hold different values for $d$ in three different ways: i) ($V_d^i = 1$, $V_d^j = 0$) or vice-versa; ii) ($V_d^i = 1$, $V_d^j = 2$) or vice-versa; iii) ($V_d^i = 0$, $V_d^j = 2$) or vice-versa. To determine $[V_d^*]_{ij}$ for $d$ in the newly induced rule $R_{ij}$, we have to first consider carefully $CF(R_i)$ and $CF(R_j)$. If $CF(R_i)>CF(R_j)$, then $[V_d^*]_{ij} = V_d^i$. If $CF(R_j)>CF(R_i)$, then $[V_d^*]_{ij} = V_d^j$. On the other hand, if $CF(R_i) = CF(R_j)$, then $[V_d^*]_{ij} = 2$. We now determine in equation (iii) how to calculate $CF^*(R_{ij})$.

$$CF^*(R_{ij}) = \begin{cases} |(CF(R_i)C_w^i - CF(R_j)C_w^j)|, & CF(R_i)C_w^i > \alpha \text{ and } CF(R_j))C_w^j > \alpha \\ CF(R_i)C_w^i, & CF(R_j)C_w^j \le \alpha \text{ and } CF(R_i)C_w^i > \alpha \\ CF(R_j)C_w^j, & CF(R_i)C_w^i \le \alpha \text{ and } CF(R_j)C_w^j > \alpha \\ 0, & CF(R_i)C_w^i \le \alpha \text{ and } CF(R_j)C_w^j \le \alpha \end{cases} \quad \text{........... (iii)}$$

Now we describe the mechanism for calculating the credibility factor for the newly induced rule $R_{ij}$ when we have knowledge sources for $R_{ij}$. In this scenario, if $[V_d^*]_{ij}$ for $d$ in $D_{ij}$, as measured from the constituent rules, is same as the truth value as given by the knowledge sources, then we calculate $exp1 = CF^\#(R_{ij}) + CF^*(R_{ij})$ where $CF^\#(R_{ij})$ is the credibility factor for $R_{ij}$ as calculated from supplying knowledge sources. If $(exp1 > 1)$, then $CF(R_{ij}) = 1$, else $CF(R_{ij}) = exp1$. But, if $[V_d^*]_{ij}$ for $d$ in $D_{ij}$ as measured from constituent rules is different from truth value ($[V_d^\#]_{ij}$) for $d$ in $D_{ij}$ as given by the external evidence, we determine $[V_d]_{ij}$ and the credibility factor for $R_{ij}$ as per the following algorithm.

Procedure Credibility_ $R_{ij}$_External($CF^\#(R_{ij})$, $CF^*(R_{ij})$)
    If ($CF^\#(R_{ij}) > CF^*(R_{ij})$) Then
        $[V_d]_{ij} = [V_d^\#]_{ij}$;
        $CF(R_{ij}) = CF^\#(R_{ij}) - CF^*(R_{ij})$;
        Elseif ($CF^\#(R_{ij}) < CF^*(R_{ij})$) Then
            $[V_d]_{ij} = [V_d^*]_{ij}$;
            $CF(R_{ij}) = CF^*(R_{ij}) - CF^\#(R_{ij})$
        Else
            $[V_d]_{ij} = 2$;
            $CF(R_{ij}) = [T_v^3]_{ij}$;
    Endif
End Credibility_ $R_{ij}$_External

The algorithm takes constant execution time.

So far we have considered only two constituent atomic rules for inducing a new rule. But as per our lattice based KB, a rule $R_{new}$ at level $i$ ($i$>2) may be constructed using a set of $i$ constituent rules $R_c$, $R_{c+1}$,…, $R_{c+i}$ ($c$>0). Suppose all the constituent rules contain $d$ as part of their decisions. We will first show how to determine $V_d$ for $d$ in $R_{new}$ using the following steps: starting from extreme left of the lattice diagram, we first take two constituent rules $R_c$ and $R_{c+1}$ from the set of $i$ rules to construct a dummy rule $R_{dummy1}$. We then check whether $CF(R_c) > CF(R_{c+1})$. If it is, then $V_d$ for $R_{dummy1}$ will be same as that of $R_c$. If $CF(R_c) < CF(R_{c+1})$, then $V_d$ for $R_{dummy1}$ will be same as that of $R_{c+1}$. If $CF(R_c) = CF(R_{c+1})$, then $V_d$ for $R_{dummy1}$ will be same as that of either $R_c$ or $R_{c+1}$. Now, we take $R_{dummy1}$ and the third constituent rule $R_{c+2}$ to construct another dummy rule $R_{dummy2}$ and carry out the same procedure as $R_{dummy1}$ for calculating $V_d$ of $d$ for $R_{dummy2}$. In this way we proceed until we obtain the rule $R_{dummy(i-1)}$ and calculate its $V_d$. Now we use the credibility factors of $R_{dummy(i-1)}$ and that of the $i^{th}$ constituent rule $R_{c+i}$ to calculate the final $V_d$ of the rule $R_{new}$.

We now show how to determine the credibility factor for $d$ in $R_{new}$. We assume that the condition part of $R_{new}$ consists of facts $f_k$, $f_{k+1}$, $f_{k+2}$,…, $f_{k+i}$ ($k$>0) associated with conditional weightages $C_w^k$, $C_w^{k+1}$, $C_w^{k+2}$ ,…, $C_w^{k+i}$ respectively. If we consider all the condition parts of constituent rules, then we would find each of the facts ($i$–1) times. Using the implicit lattice knowledge, we calculate the credibility factor for $R_{new}$ as follows. For this, we consider that a sub-set of constituent rules $S_k \subset \{R_c, R_{c+1},\ldots, R_{c+i}\}$ that hold fact $f_k$ in the condition parts. Similarly, $S_{k+1} \subset \{R_c, R_{c+1},\ldots, R_{c+i}$ $\}$ holds fact $f_{k+1}$ and so on. Obviously, $S_k \cap S_{k+1} \cap \ldots \cap S_{k+i} \neq \phi$. Now consider $S_k$. We check $V_d$ of each of rules belonging to $S_k$. Suppose there are $m_k$ rules whose $V_d$ match with that of $R_{new}$, and $n_k = (|S_k| - m_k)$ rules whose $V_d$ do not match with that of $R_{new}$. We define $CF_k = \sum_{a=0}^{m_k} CF(S_k^a) - \sum_{b=0}^{n_k} CF(S_k^b)$ , where $S_k^a$ ($0 \leq a \leq m_k$) represents a rule belonging to $S_k$ with same $V_d$ as $R_{new}$, and $S_k^b$ ($0 \leq b \leq n_k$) represents a rule belonging to $S_k$ with different $V_d$ compared to that of $R_{new}$. Then, we consider $S_{k+1}$. We check $V_d$ of each of rules belonging to $S_{k+1}$. Suppose there are $m_{k+1}$ rules whose $V_d$ match with that of $R_{new}$, and $n_{k+1}$ rules whose $V_d$ do not match with that of $R_{new}$. We define $CF_{k+1} = \sum_{a=0}^{m_{k+1}} CF(S_{k+1}^a) - \sum_{b=0}^{n_{k+1}} CF(S_{k+1}^b)$ , where $S_{k+1}^a$ ($0 \leq a \leq m_{k+1}$) represents a rule belonging to $S_k$ with same $V_d$ as $R_{new}$, and $S_{k+1}^b$ ($0 \leq b \leq n_{k+1}$) represents a rule belonging to $S_k$ with different $V_d$ compared to that of $R_{new}$. In this way we find $CF_{k+2}$, $CF_{k+3}$, …, $CF_{k+i}$ corresponding to $S_{k+2}$, $S_{k+3}$, …, $S_{k+I}$ respectively. So, the credibility factor for $R_{new}$ is calculated using equation (iv).

$$CF(R_{new}) = \frac{1}{(i-1)}(CF_k C_w^k + CF_{k+1} C_w^{k+1} + \cdots + CF_{k+i} C_w^{k+i}), \text{if } CF_j C_w^j > \alpha \text{ for } k \leq j \leq k+i \quad \text{……….. (iv)}$$

If $CF(R_j)C_w^j \leq \alpha$ for some $j$, then the value of $CF(R_j)C_w^j$ = 0. If for all $j$, we get $CF(R_j)C_w^j \leq \alpha$, then $CF(R_{new})$ = 0. If the R.H.S of equation (iv) is greater than 1, then $CF(R_{new})$ = 1.

One thing to be mentioned here is that if a rule $R_{new}$ is induced from two constituent rules $R_{con1}$ and $R_{con2}$, and $d$ is available in only one constituent rule $R_{con2}$ with the conditional weightage $C_w$ for fact $f_{con2}$, then $d$ would also be available in the decision part of rule $R_{new}$ with same truth value as $R_{con2}$ if $CF(R_{con2})C_w > \alpha$. Otherwise, $d$ will not be present as part of the decision for $R_{new}$.

4.1.2 Data structure for the proposed lattice

We have proposed a new lattice based data structure which would be followed to develop the knowledge base for a medical expert system. The lattice based KB is a multi-level non-linear bottom-up data structure with each level containing a number of nodes to hold the lattice elements. Each level can be visualized as a special type of node called a '*Lattice_Level* node' which contains four fields: a) a pointer to the immediate Successor *Lattice_Level* node, b) the current level, c) total number of lattice elements in the current level, d) array of pointers to hold all the lattice elements in the current level. The top-most *Lattice_Level* node has an empty immediate successor field. Each lattice element is represented by another type of node called '*Lattice_Element* node' which contains five fields: a) a boolean label for the node, b) the 'condition' part $C$ of a rule, c) the 'decision' part $D$ of the same rule, d) a pointer to point the first *Lattice_Element* node of the predecessor list which is composed of related *Lattice_Element* nodes residing at the immediate predecessor level, e) a pointer to point the first *Lattice_Element* node of the successor list which is composed of related *Lattice_Element* nodes residing at the immediate successor level. So, every *Lattice_Element* node corresponds to a rule of the form $C \rightarrow D$.

Label of a *Lattice_Element* node is a means for easy identification and easy reference to a *Lattice_Element* node in KB. A label is a boolean string of $n$ bits, if the lattice is of order $n$. The *Lattice_Element* node present at level 0 is a boolean string of $n$ 0s. The *Lattice_Element* node present at level $n$ is a boolean string of $n$ 1s. As a lattice of order $n$ has $n$ *Lattice_Element* nodes corresponding to $n$ atomic rules at level 1, the label of each of the nodes would contain only one 1 and ($n$–1) 0s. The label of the first *Lattice_Element* node from the left hand side of the lattice would be represented as $0^{n-1}1$. The label of the next *Lattice_Element* node would be represented as $0^{n-2}10$, which is obtained by left-shifting the previous label by 1 bit. If $l_i$ is a *Lattice_Element* node corresponding to $i^{\text{th}}$ atomic rule at level 1, the label of $l_i$ is $0^{(n-(i+1))}10^i$, which can be obtained by left-shifting the label of $l_{i-1}$. In this way we proceed until we find the label of the right-most *Lattice_Element* node at level 1 as $10^{n-1}$. The *Lattice_Element* nodes at level 2 would have exactly two 1s in their labels. More elaborately, the *Lattice_Element* node with label $0^{n-1}1$ and the *Lattice_Element* node with label $0^{n-2}10$ would form the *Lattice_Element* node at level 2 whose label is $0^{n-2}11$, i.e., the label $0^{n-2}11$ is the 'Boolean OR' combination of $0^{n-1}1$ and $0^{n-2}10$ . So, in general, as a *Lattice_Element* node at level 2 is constructed using two *Lattice_Element* nodes at level 1, the label of each *Lattice_Element* node is obtained by ORing the labels of its constituent rules. Similarly, label of every *Lattice_Element* node at level $i$ ($1\leq i \leq n$) is obtained by ORing the labels of all its constituent *Lattice_Element* nodes at $(i–1)^{th}$ level of the lattice.

The lattice based data structure can be visualized as a bi-directed graph $G = (V, E)$, where $V$ consists of two types of nodes: *Lattice_Level* nodes $L_v$, and *Lattice_Element* nodes $L_n$. In a lattice based data structure of order $n$, there are ($n$+1) $L_v$ nodes and $2^n$ $L_n$ nodes. For each $L_n$-type node $L_n^j$ at $i^{\text{th}}$ level ($0 < i \leq n$ and $1 \leq j \leq {}^n\text{C}_i$), we create an adjacency list called 'Predecessor_Adjacency_List' to hold all the $L_n$ nodes residing at the immediate predecessor level ($i$–1), that are attached to the node $L_n^j$. The fourth field of a *Lattice_Element* node holds the address of the first node of the corresponding Predecessor_Adjacency_List. Again, for each $L_n$ node $L_n^j$ at $i^{\text{th}}$ level ($0 \leq i < n$), we create another adjacency list called 'Successor_Adjacency_List' to hold all the $L_n$ nodes, at the immediate successor level ($i$+1), that are attached to the node $L_n^j$. The fifth field of a *Lattice_Element* node holds the address of the first node of the corresponding Successor_Adjacency_List. $E$ consists of arcs between $L_n$ nodes of two adjacent levels. No edges exist among the $L_n$ nodes at the same level.

In a structured and concrete way, we define the lattice based KB of order $n$ ($n$ is the number of atomic rules) with ($n$+1) levels as a bottom-up non-linear data structure containing

i) a *head* node, which is different from a *Lattice_Level* node or a *Lattice_Element* node, to hold the total number of levels in the lattice and a pointer to hold the address of the *Lattice_Level* node at the $0^{th}$ level.

ii) a *Lattice_Level* node at the $0^{th}$ level to hold only a root *Lattice_Element* node whose second, third and fourth fields are empty. The *Lattice_Element* node has a *Successor_Adjacency_List* of $n$ *Lattice_Element* nodes at $1^{st}$ level.

iii) a *Lattice_Level* node at the $n^{th}$ level to hold only an end/terminal *Lattice_Element* node whose fifth field is empty. This *Lattice_Element* node has a Predecessor_Adjacency_List of $n$ *Lattice_Element* nodes at $(n-1)^{th}$ level.

iv) a *Lattice_Level* node at the $l^{th}$ level ($0 < l < (n+1)$) to hold ${}^nC_l$ *Lattice_Element* nodes with all the fields of each *Lattice_Element* node being non-empty. Each *Lattice_Element* node has a Successor_Adjacency_List of ($n$–$l$) *Lattice_Element* nodes and a Predecessor_Adjacency_List of $l$ *Lattice_Element* nodes.

We now provide the structural definitions of different types of nodes used for our design using C language like constructs.

a. Structural Definition of *head* node

```
struct head
{
int count_level;                    // count_level indicates the total no. of levels present in the lattice
```

struct *Lattice_Level_Node* **ptr*; // *ptr* points to the *Lattice_Level* node at $0^{th}$ level
}

b. Structural Definition of a *Lattice_Level* node

struct *Lattice_Level_Node*
{
struct *Lattice_Level_Node* **next_level*; //*next_level* is the immediate successor *Lattice_Level* node
int *level*; //*level* indicates the current level for the *Lattice_Level* node under consideration
int *total_node_level*; /* *total_node_level* indicates the total no. of *Lattice_Element* nodes present at the current level, which is basically ${}^{n}C_{l}$ where $l$ is a level ($0 \leq l \leq n$) */
struct *Lattice_Element_Node* **arr*[]; /**arr* is an array of pointers to hold all the *Lattice_Element* nodes present at the current level*/
}

c. Structural Definition of a *Lattice_Element* node

struct *Lattice_Element_Node*
{
bool *label*[]; //*label* of a *Lattice_Element* node is represented by a boolean string
char *condition*[]; //*condition* of a *Lattice_Element_Node* is basically a character string
struct *Decision* **decision*; //*decision* is instantiation of a *Decision* structure
struct *Node* **predecessor*; //*predecessor* is the pointer to the Predecessor_Adjacency_List
struct *Node* **successor*; //*successor* is the pointer to the Successor_Adjacency_List
}

struct *Decision*
{
struct *Decs* **D*[] ; //*Decs* structure is used for formation of the pair $<d, V_d>$
int *tot_decs*; //*tot_decs* keeps track of the total number of decisions for a *Lattice_Element_Node*
}

struct *Decs*
{
char *disease*[]; //*disease* of a rule is a character string
float *Cond_Weight*[*N*]; /**Cond_Weight*[*i*] [$1 \leq i \leq N$] corresponds to the conditional weight of $i^{th}$ fact for a decision, with *N* being the total number of constituent facts */
float $T_v^1$;
float $T_v^2$;
float $T_v^3$;
int *truth_value*; // *truth_value* stands for $V_d$ of *d*
float *CF*; //*CF* represents the Credibility_Factor
}

struct *Node*
{
bool *label*[]; // *label* of a *Lattice_Element* node
struct *Node* *ptr; // *ptr* points to the next node in predecessor or successor adjacency list
}

In the following section, we give an algorithm for the knowledge base construction. We have used mixtures of natural languages and C language-like constructs for describing the algorithms.

4.1.3 Design of an algorithm for knowledge base construction

Procedure *Build_KB* (*n*) // *n* is the no. of atomic rules

```
struct head *h = NIL;
struct Lattice_Level_Node *l_v = NIL, *prev = NIL;
struct Lattice_Element_Node *l_n = NIL;

/* Creation of Lattice_Level node at level 0 */
count_level [h] = n+1;  //h keeps track the total number of levels in KB
level = 0;
l_v = Create_Level_Node (prev, level)  //The procedure Create_Level_Node generates a Lattice_Level node
For i = 0 to total_node_level [l_v]–1 Do
      arr[l_v]_i = Create_Lattice_Node (l_v);  /* The procedure Create_Lattice_Node generates a
                                                          Lattice_Element node */
Endfor
ptr[h]  = l_v;

/* Creation of all the Lattice_Level nodes and corresponding Lattice_Element nodes */
level = level + 1;
prev = l_v;
While (level ≠ count_level)
      l_v = Create_Level_Node (prev, level);
      For i = 0 to total_node_level [l_v]–1 Do
             arr[l_v]_i = Create_Lattice_Node (l_v);
      Endfor
      prev ← l_v;
      level ← level + 1;
Endwhile

/* Creation of adjacency lists of predecessor and successor nodes of each Lattice_Element node */
l_v ← ptr[h];
While (l_v ≠ NIL)
      For i = 0 to total_node_level [l_v]–1 Do
             predecessor [arr[l_v]_i] ← Create_Predecessor_List (arr[l_v]_i, level[l_v]); /* The procedure
             Create_Predecessor_List creates Predecessor_Adjacency_List for a Lattice_Element_Node */
             successor [arr[l_v]_i] ← Create_Successor_List (arr[l_v]_i, level[l_v]); /* The procedure
             Create_Successor_List creates Successor_Adjacency_List for a Lattice_Element_Node */
       Endfor
       l_v ← next_level[l_v];
 Endwhile

 /*Returning the starting pointer of KB*/
 Return h;
End Build_KB
```

The procedure *Create_Level_Node*(*prev*,*level*) creates a new *Lattice_Level_Node* $l_v$ which resides as the next level of *prev*. The *next_level* field of $l_v$ is NIL and the total number of *Lattice_Element* nodes at *level* is ${}^{n}C_{level}$. The procedure *Create_Lattice_Node*($l_v$) creates a new *Lattice_Element* node $l_n$ attached to $l_v$. The label of $l_n$ is generated as per the process described earlier. Please refer to the Appendix A for detailed procedures.

*Complexity analysis of Build_KB procedure*

As per our proposed design, $|V| = 2^n$. So, $n = \log_2 |V|$. Suppose, the time taken by the procedure *Create_Level_Node* is $T_1$ and that of the procedure *Create_Lattice_Node* is $T_2$, where $T_1 \neq T_2$. The *Create_Predecessor_List* and *Create_Successor_List* procedures take $O(\log_2|V|)$ time each to execute. We assume that the (floor($n$/2))$^{th}$ level holds $N$ number of *Lattice_Element* nodes, where $N = {}^{n}C_{n/2}$. In a lattice of order $n$, $N$ is the maximum number of *Lattice_Element* nodes that a *Lattice_Level* node can hold. Therefore the time complexity of *Build_KB* algorithm is $O(\log_2|V|(T_1+NT_2))$. The algorithm is a polynomial-time algorithm. As the proposed design is basically a graph represented using adjacency list, the space complexity of the algorithm is $O(|V|+|E|)$.

As our proposed design supports the construction of knowledge base based on the availability of domain knowledge at a particular time, the lattice based KB should have some scopes for insertion of additional knowledge, modification or deletion of some existing knowledge.

*Insertion of a new fact f into the existing lattice-based KB*

Insertion of a new fact can only be done at a new *Lattice_Element_Node* placed at the right-most end of the level 1 of the existing lattice KB. The label of the new *Lattice_Element_Node* is $10^n$. With the insertion, the length of each label of each existing *Lattice_Element_Node* is increased by one. Except level 0, each *Lattice_Level_Node* $l_v$ at upper levels would now hold $^{(n+1)}C_{(\text{level}[lv])}$ *Lattice_Element* nodes. So, at each level $(^{(n+1)}C_{(\text{level}[lv])} - {}^{n}C_{(\text{level}[lv])})$ new *Lattice_Element_Node*s would be created at the right. A new level on the top of the existing KB would be created and this new level will have only one *Lattice_Element* node. Now, we have to modify the Predecessor_Adjacency_List and the Successor_Adjacency_List of every *Lattice_Element_Node* in the KB so that the updated KB follows a lattice structure of order ($n$+1).

*Deletion of fact f from KB*

This operation deletes a fact *f* from all the *Lattice_Element* nodes which hold *f* in their condition parts. If a *Lattice_Element_Node* is deleted at level 1, then all its successors upto the topmost level of KB would be deleted. The topmost level would also be deleted. Now the lattice is of order ($n$ – 1). Except level 0, each *Lattice_Level_Node* $l_v$ at upper levels would now hold $^{(n-1)}C_{(\text{level}[lv])}$ *Lattice_Element_Node*s. So, modifications should be made at the Predecessor_Adjcency_List and the Successor_Adjacency_List of every *Lattice_Element_Node* that is present in KB after the deletion of nodes. We have to now make some modifications in the labels of existing *Lattice_Element* nodes. For this, we check whether there is any ON bit (bit 1) present in the left substring with respect to the location of *f* (obviously the location of *f* contains a 0 bit) in the string of a label of a *Lattice_Element* node. If so, then there would be no change in the label. If there is any ON bit present in the right substring with respect to the location of *f* in the string, then we right-shift the ON bits by one place. Finally, we have to cut down the left-most 0 of every label.

*Modification of a Lattice_Element_Node in KB*

Modification of a *Lattice_Element_Node* means changing its condition or decision part. Change in condition part can be performed only at 1*st* level. This change would be reflected into the condition parts of its successors at upper levels of the lattice. Change in condition part may or may not result in changes in decision part. If there is change in the decision part, then either a set of new decisions may be added to the existing decision part, or a set of existing decisions is replaced by a set of new decisions, or a set of existing decisions may be deleted. Changing a decision part includes modification of conditional weightages and credibility factors.

## 4.2 Handling inconsistent knowledge using rough set

The lattice-based KB may contain many inconsistent rules in which the condition parts imply inconsistent decisions. An inconsistent decision holds an inconclusive truth-value for a disease. Presence of inconsistent rules in a medical knowledge base is an obvious disadvantage as the medical expert system would not be able to infer diagnostic and therapeutic decisions accurately and reliably using these kinds of rules. With the aim of producing a dependable and reliable medical expert system, we use rough set theory to deal with the inconsistent rules.

### 4.2.1 Approximation set generation for inconsistent knowledge in KB

We consider our proposed lattice-based KB as an information system which may be expressed as $S_r = (U_r, C_r, D_r)$, where $U_r$ is a finite set of *Lattice_Element* nodes with distinct labels, $C_r$ is a set of facts under consideration and $D_r$ is the set of decisions. As we have said earlier, let us consider a *Lattice_Element* node $l_n$ in KB which corresponds to an individual rule $R_{l1}$: $C_{l1} \rightarrow D_{l1}$, which may not be optimal. There may be another rule in KB $R_{l2}$: $C_{l2} \rightarrow D_{l2}$, where $C_{l2} \subset C_{l1}$ and $D_{l1} = D_{l2}$. As every *Lattice_Element* node has a label of a boolean string, we can say that every attribute/fact $f \in C_r$ is associated with a value of either 0 or 1. Every attribute $d \in D_r$ is associated with a set of values from $V_d$. As every label is distinct for every rule in KB, the indiscernibility relation $I(C_r)$ would produce $|U_r|$ singleton elementary sets. As there are $|U_r|$ elementary sets, the reduct of the original set of attributes in KB is same as the set $C_r$.

We have seen previously that there may be many *Lattice_Element* nodes in the KB, which include one or more inconsistent decisions in their 'decision' parts. The easy method to identify inconsistent decision is to check whether $V_d$ is equal to 2. If so, then the decision is inconsistent. We define two concepts, where the first concept is called $\text{Concept}_1$ and holds the elementary sets associated with $V_d = 1$ for a particular decision *d*, and second concept is termed as $\text{Concept}_2$ which holds the elementary sets associated with $V_d = 0$ also for the decision *d*. As an inconclusive decision means that the disease may or may not be present, all the decisions of the rules with $V_d = 2$ must be included in both the concepts. Obviously, $\text{Concept}_1$ and $\text{Concept}_2$ contain both the consistent and inconsistent rules. We now calculate the lower and upper approximations of both the concept using the procedure *Approximation_Generation_KB*. The lower and upper approximation sets generated in this way also support the basic properties as stated by (Komorowski et al. 1999).

```
Procedure Approximation_Generation_KB (h, d) //h is the starting address of the lattice-based KB
     int l1 = 0, l2 = 0, l3 = 0;
    ArrD = Search_Decision_KB(h, d); /*ArrD holds the Lattice_Element nodes which have d as one
                                      of the decisions */
    count = Size_of(ArrD);           /* count is a temporary variable which holds the size of ArrD */

    /* Generation of upper and lower approximations for Concept1 */
    For i = 0 to count – 1 Do
        If Vd == 1 OR Vd == 2 Then
            Upper_Approximation1[l1] = ArrDi ;      //Upper Approximation for Concept1
            l1 = l1 + 1;
        Endif
        If Vd == 1 Then
            Lower_Approximation1[l2] = ArrDi ;      //Lower Approximation for Concept1
            l2 ← l2 + 1;
        Endif
        If Vd == 2 Then
            Boundary_Region1[l3] = ArrDi ;          //Boundary region for Concept1
            l3 ← l3 + 1;
        Endif
    Endfor

    /* Generation of upper and lower aproximations for Concept2 */
   l1 = 0, l2 = 0, l3 = 0;
   For i = 0 to count – 1 Do
        If Vd == 0 OR Vd == 2 Then
            Upper_Approximation2[l1] = ArrDi ;      //Upper Approximation for Concept2
            l1 ← l1 + 1;
        Endif
        If Vd == 0 Then
            Lower_Approximation2[l2] = ArrDi ;      //Lower Approximation for Concept2
            l2 ← l2 + 1;
        Endif
        If Vd == 2 Then
            Boundary_Region2[l3] = ArrDi;           //Boundary region for Concept2
            l3 ← l3 + 1;
        Endif
    Endfor
End Approximation_Generation_KB
```

The time complexity of the *Approximation_Generation_KB* algorithm is $O(\log_2|V|)$.

Now, we define how the sets of approximations can be modified when there are some changes in the KB. Our lattice-based design allows modification in approximation generation without generating the approximated sets again from scratch, thereby reducing the modification time. Modification of approximation generation can be done when a set of existing decisions are added to some other *Lattice_Element* nodes in KB, when some existing decisions are deleted from KB, and when truth values of some decisions are changed. Please refer to the Appendix B for detailed algorithm.

### 4.2.2 Optimal rule generation from KB

As per definitions of approximations of rough set, from the lower approximation of a concept, we find a set of certain rules. The rules generated from the upper approximation of a concept are called possible rules. The rules that are generated from the boundary region of a concept are called uncertain rules. Our proposed design also produces certain rules from the lower approximation sets generated for all the diseases present in KB. We also can get possible and uncertain rules from the upper approximation sets and boundary region sets respectively obtained from the lattice based information system.

As our rough set based lattice knowledge representation technique generates an exhaustive set of certain, possible, and uncertain rules given a set of facts, we may find many rules with common condition parts. Also, we may find a rule $R_a$ with condition $C_a$ and decision $D$, and another rule $R_b$ with condition $C_b$ and decision $D$ where $C_a \subset C_b$. This would result in poor performance during the time of forward reasoning on the condition $C_a$, as both the rules would be fired simultaneously. To avoid this kind of redundancy in the generated rule set, we propose an efficient rule minimization technique. The rules in the minimized rule set would hold only one decision in their decision parts.

*Minimization of rules*

We can minimize the rules using the concept of minimization of boolean functions in digital logic. We know that a boolean function can be expressed as a sum of minterms or product of maxterms. The process of simplifying the algebraic expression of a Boolean function is called function minimization. The importance of minimization lies in reduction of cost and complexity of associated circuit. In this paper, we want to apply the function minimization technique to have minimum number of conditions against a decision. This would obviously reduce the inferencing time for a medical expert system. It is very much easy for the proposed design to apply function minimization procedure as we have labeled the *Lattice_Element* nodes using boolean strings.

Suppose, for a disease $d$ to be present i.e. ($V_d$ = 1), a set of rules $S_R = \{R_i \mid R_i$ is a rule and $1 \le i \le n\}$ have been found from a lattice-based KB of order $n$. Our assumption in this context is that if $R_i$ is there, $R_{i+1}$ may not be present in $S_R$. If $R_{k1} \in S_R$ and $R_{k2} \in S_R$ where $1 \le k1$, $k2 \le n$, we can generate a minimized rule as (Condition[$R_{k1}$] + Condition[$R_{k2}$]) $\rightarrow$ ($d$,1). This kind of consideration is similar to expressing the condition parts of elements of set $S_R$ as sum of minterms. As each rule corresponds to a particular *Lattice_Element* node in the KB, the minterms are basically the labels of that particular nodes. Their sum is used to find out the minimum number of facts conditions. The minimized rule may end up with one or more conditions with or without any 'OR' operator among them. One thing to be mentioned here is that *K*-map procedure for boolean function minimization would not be applicable here as we have binary-coded the *Lattice_Element* nodes without using the actual binarization process.

As an example, we consider a lattice-based KB of order 4 with four different conditions $c_1$, $c_2$, $c_3$ and $c_4$ for four atomic rules available at level 1. Suppose a particular decision ($d$,1) occurs for *Lattice_Element* nodes with labels 1000, 1001, 1101 and 1100. So, we define the sum of minterms as ($c_4c_3'c_2'c_1' + c_4c_3'c_2'c_1 + c_4c_3c_2'c_1 + c_4c_3c_2'c_1'$) which would be minimized as $c_4c_2'$. So, the minimized rule is: $c_4c_2' \rightarrow (d, 1)$. The implication is that if a patient has symptom $c_4$ but he/she has no symptom of $c_2$, then we would infer that the disease $d$ may be present for the patient. It is to be noted here that associated with $c_4$, a patient may have the other symptoms $c_1$ or/and $c_3$, but he/she does not have at all the symptom c2. Let us take another example with the same lattice-based KB. We consider that another decision ($d_1$, 1) occurs for *Lattice_Element* nodes with labels 1100, 1101, 1001, 1111, 1011, 1110, and 1010. In this case, we define the sum of minterms as ($c_4c_3c_2'c_1' + c_4c_3c_2'c_1 + c_4c_3'c_2'c_1 + c_4c_3c_2c_1 + c_4c_3'c_2c_1 + c_4c_3c_2c_1' + c_4c_3'c_2c_1'$) which would be minimized as [$c_4(c_3 + c_2 + c_1)$]. So, the minimized rule is: $c_4(c_3 + c_2 + c_1) \rightarrow (d_1,1)$. The implication is that if a patient has symptom $c_4$ and associated with this, he/she has any other symptom from the set $\{c_3, c_2, c_1\}$, then we can infer that the disease $d_1$ may be present for the patient.

4.2.3 Defining certainty and coverage factors for generated optimal rules

In classical rough set theory, while calculating the certainty factor and coverage factor of a rule, we use all the reliable and unreliable data sources. As reliability is an important concern in medical expert systems, we would use only the reliable sources for calculating certainty and coverage factors of a rule generated from the lattice-based KB. We re-define the 'Support' of an optimal rule keeping in mind its credibility factor.

**Definition 1**: **Support** of an optimal rule generated from lattice based KB (Ȿ)

Suppose $R$: $C \rightarrow D$ is an optimal rule, which has been obtained by minimizing a set of rules namely $R_{i1}$: $C_{i1} \rightarrow D$, $R_{i2}$: $C_{i2} \rightarrow D$, …, $R_{ik}$: $C_{ik} \rightarrow D$ ($i$,$k$>0); i.e. $C_{i1} \cap C_{i2} \cap \ldots \cap C_{ik} = C$. We define the support (Ȿ) of each rule belonging to the set $\{R_{i1}, R_{i2}, \ldots, R_{ik}\}$ as

Ȿ($R_{ij}$) = $CF(R_{ij})$, where $1 \leq j \leq k$

Here Ȿ($R_{ij}$) is called the 'fractional support' of $R_{ij}$ and only takes into account the reliable knowledge sources, discarding the unreliable or unacceptable knowledge sources.
Now, we calculate the fractional support of the optimal rule $R$ as follows:

Ȿ($R$) = $\sum_{n=i1}^{ik} CF(R_n)$

**Definition 2: Strength** of an optimal rule generated from lattice based KB (ƻ)

We consider that, there exist $l$ (>0) rules such as $R_1$: $C_1 \rightarrow D$, $R_2$: $C_2 \rightarrow D$, …, $R_l$: $C_l \rightarrow D$, where $\{R_{i1}, R_{i2}, \ldots, R_{ik}\} \subseteq \{R_1, R_2, \ldots, R_l\}$. Please note that $V_d$ of all the rules belonging to the set $\{R_1, R_2, \ldots, R_l\}$ might not be same. From the set $\{R_1, R_2, \ldots, R_l\}$, we calculate Ꝑ$_D$ = $\sum_{m=1}^{l} CF(R_m)$. Now, we define the strength (ƻ) of each rule $R_j$ belonging to the set $\{R_1, R_2, \ldots, R_l\}$ as

ƻ($R_j$) = Ȿ($R_j$)/Ꝑ$_D$, where $1 \leq j \leq l$

Here, ƻ($R_j$) is called the 'reliability strength' of $R_j$. Now, we calculate the reliability strength of the optimal rule $R$ as follows:

ƻ($R$) = Ȿ($R$)/Ꝑ$_D$

If $\{R_{i1}, R_{i2}, \ldots, R_{ik}\}=\{R_1, R_2, \ldots, R_l\}$, then the strength of the rule $R$ would be 1. So, the strength of a rule will remain between 0 and 1.

**Definition 3: Certainty Factor** of an optimal rule generated from lattice based KB (Ʒ)

As per our preliminary assumption, the set of rules $\{R_{i1}, R_{i2}, \ldots, R_{ik}\}$ hold $C$ in their condition parts with $V_d = x$, ($x$ =1 or 0). We first determine the set of rules $\{R_{l1}, R_{l2}, \ldots, R_{lm}\}$from the set $\{R_1, R_2, \ldots, R_l\}$where $V_d = x$. Obviously, $\{R_{i1}, R_{i2}, \ldots, R_{ik}\} \subseteq \{R_{l1}, R_{l2}, \ldots, R_{lm}\}$. We determine ¥$_C$ = $\sum_{d=i1}^{ik} CF(R_d)$. We define the certainty factor (Ʒ) of each rule $R_k$ belonging to the set $\{R_{i1}, R_{i2}, \ldots, R_{ik}\}$ as follows:

Ʒ($R_k$) = Ȿ($R_k$)/¥$_C$ = ƻ($R_k$) / $\Pi(C)$, where $\Pi(C)$ = ¥$_C$/Ꝑ$_D$
So, Ʒ($R$) = Ȿ($R$)/¥$_C$ = ƻ($R$) / $\Pi(C)$, where $\Pi(C)$ = ¥$_C$/Ꝑ$_D$

Now, let's forget about the set $\{R_{i1}, R_{i2}, \ldots, R_{ik}\}$. Suppose a set of rules $\{R_{j1}, R_{j2}, \ldots, R_{jg}\}$ ($j$, $g$>0) hold $C$ in their condition parts with $V_d$ = 2. We then also search for the set of rules $\{R_{h1}, R_{h2}, \ldots, R_{hp}\}$ ($h$, $p$>0) from the set $\{R_1, R_2, \ldots, R_l\}$where $V_d$ = 2. In this case, we determine ¥$_C$ = $\sum_{d=j1}^{jg} CF(R_d)$. So, the certainty factor (Ʒ) of each rule $R_j$ belonging to the set $\{R_{j1}, R_{j2}, \ldots, R_{jg}\}$ as follows:

$$ʒ(R_j) = ½\ [Ş(R_j)/¥_C] = ½\ [ƻ(R_j)\ /\ \Pi(C)],\ \text{where}\ \Pi(C) = ¥_C/₱_D$$
$$\text{So, } ʒ(R) = ½\ [Ş(R)/¥_C] = ½\ [ƻ(R)\ /\ \Pi(C)],\ \text{where}\ \Pi(C) = ¥_C/₱_D$$

Here the fraction '½' comes because, for each rule $R_j$ we get two conclusions in the consistent rule set, one with $V_d = 1$, and another with $V_d = 0$.

If $ʒ(R) = 1$, $R$ is called a certain rule.

**Definition 4: Coverage Factor** of an optimal rule generated from lattice based KB (Ȝ)

With the same considerations as in Definition 3, we determine $℘_{(Vd = x)} = \sum_{d=l1}^{lm} CF(R_d)$, where $x \neq 2$. We define the coverage factor (Ȝ) of each rule $R_k$ belonging to the set $\{R_{i1}, R_{i2}, \ldots, R_{ik}\}$ as follows:

$$Ȝ(R_k) = Ş(R_k)/℘_{(Vd = x)} = ƻ(R_k)\ /\ \Pi(V_d = x),\ \text{where}\ \Pi(V_d = x) = ℘_{(Vd = x)}/₱_D$$
$$\text{So, } Ȝ(R) = Ş(R)/℘_{(Vd = x)} = ƻ(R)\ /\ \Pi(V_d = x),\ \text{where}\ \Pi(V_d = x) = ℘_{(Vd = x)}/₱_D$$

Now, let's forget about the set $\{R_{i1}, R_{i2}, \ldots, R_{ik}\}$. Suppose a set of rules $\{R_{j1}, R_{j2}, \ldots, R_{jg}\}$ ($j$, $g$>0) hold $C$ in their condition parts with $V_d = 2$. In this case, we determine $℘_{(Vd = 2)} = \sum_{d=h1}^{hp} CF(R_d)$. So, the coverage factor (Ȝ) of each rule $R_j$ belonging to the set $\{R_{j1}, R_{j2}, \ldots, R_{jg}\}$ as follows:

$$Ȝ(R_j) = ½\ [Ş(R_j)/℘_{(Vd = 2)}] = ½\ [ƻ(R_j)/\ \Pi(V_d = 2)],\ \text{where}\ \Pi(V_d = 2) = ℘_{(Vd = 2)}/₱_D$$
$$\text{So, } Ȝ(R) = ½\ [Ş(R)/℘_{(Vd = 2)}] = ½[ƻ(R)/\ \Pi(V_d = 2)],\ \text{where}\ \Pi(V_d = 2) = ℘_{(Vd = 2)}/₱_D$$

The coverage factor gives the conditional probability of reasons for a given decision.

#### 4.2.4 Probabilistic properties of the minimized rule set

We know that any conventional decision table holds some basic probabilistic properties of rough set (Pawlak (2002)). We will now prove that the optimal decision rule set generated from lattice-based KB also holds some probabilistic properties.

**Property 1:** $\sum_{i=w1}^{wK} ʒ(R_i) = 1$

**Proof:** Suppose, we have $K$ (>0) rules namely $R_{w1}, R_{w2}, \ldots, R_{wK}$ with conditions $C_{w1}, C_{w2}, \ldots, C_{wK}$ respectively and decision $D$. Please note that $V_d$ of all the rules might not be same. Consider that $C_{w1} \cap C_{w2} \cap \ldots \cap C_{wK} = C$. Obviously, $¥_C = \sum_{d=w1}^{wK} CF(R_d)$. We now calculate the following:

$$\begin{aligned}\sum_{i=w1}^{wK} ʒ(R_i) &= Ş(R_{w1})/¥_C + Ş(R_{w2})/¥_C + \ldots + Ş(R_{wK})/¥_C \\ &= [Ş(R_{w1}) + Ş(R_{w2}) + \ldots + Ş(R_{wK})]/¥_C \\ &= \sum_{d=w1}^{wK} CF(R_d)/¥_C = 1\end{aligned}$$

**Property 2:** $\sum_{m=x1}^{xJ} Ȝ(R_m) = 1$

**Proof:** Suppose, we have $J$ (>0) rules namely $R_{x1}, R_{x2}, \ldots, R_{xJ}$ with decision $D$ where $V_d = x$ for all the rules ($x$ may be either 0 or 1). Obviously, $℘_{(V_d = x)} = \sum_{d=x1}^{xJ} CF(R_d)$. We now calculate the following:

$$\begin{aligned}\sum_{m=x1}^{xJ} Ȝ(R_m) &= Ş(R_{x1})/℘_{(Vd = x)} + Ş(R_{x2})/℘_{(Vd = x)} + \ldots + Ş(R_{xJ})/℘_{(Vd = x)} \\ &= [Ş(R_{x1}) + Ş(R_{x2}) + \ldots + Ş(R_{xJ})]/℘_{(Vd = x)} \\ &= \sum_{d=x1}^{xJ} CF(R_d)/℘_{(Vd = x)} = 1\end{aligned}$$

**Property 3:** $(\sum_{i=w1}^{wK} ʒ(R_i)).\Pi(C) = \sum_{i=w1}^{wK} ƻ(R_i)$

**Proof:** When we consider the set {$R_{w1}$, $R_{w2}$,…, $R_{wK}$} as per our preliminary assumption, $\sum_{i=w1}^{wK}$ ʒ($R_i$) = 1 according to property 1 and $\Pi(C)$ = ¥$_C$/Ᵽ$_D$. Also, ₴($R_i$) =Ȿ($R_i$)/Ᵽ$_D$. So, $\sum_{i=w1}^{wK}$ ₴($R_i$) =¥$_C$/Ᵽ$_D$. So, it is proved that L.H.S = R.H.S.

**Property 4:** ($\sum_{m=x1}^{xJ}$ Ʒ($R_m$)).$\Pi(V_d = x) = \sum_{m=x1}^{xJ}$ ₴($R_m$)

**Proof:** When we consider the set {$R_{x1}$, $R_{x2}$,…, $R_{xJ}$} as per our preliminary assumption, $\sum_{m=x1}^{xJ}$ Ʒ($R_m$) = 1 according to property 2 and $\Pi(V_d = x)$ =℘$_{(Vd = x)}$ /Ᵽ$_D$. Also, ₴($R_m$) =Ȿ($R_m$)/Ᵽ$_D$. So, ℘$_{(Vd = x)}$ = $\sum_{m=x1}^{xJ} CF(R_m)$. So, it is proved that L.H.S = R.H.S.

**Property 5:** ∀$R_i$ ∈ {$R_{w1}$, $R_{w2}$,…, $R_{wK}$}, ʒ($R_i$) = ₴($R_i$). / ($\sum_{i=w1}^{wK}$ ₴($R_i$))

**Proof:** We already know from definition 2 that ₴($R_i$) =Ȿ($R_i$)/Ᵽ$_D$. ($\sum_{i=w1}^{wK}$ ₴($R_i$)) = [Ȿ($R_{w1}$) + Ȿ($R_{w2}$) + …+Ȿ($R_{wK}$)] /Ᵽ$_D$ = ¥$_C$/Ᵽ$_D$. So, ₴($R_i$) / [Ȿ($R_{w1}$) + Ȿ($R_{w2}$) + …+Ȿ($R_{wK}$)] = Ȿ($R_i$)/¥$_C$ = ʒ($R_i$).

**Property 6:** ∀$R_j$ ∈ {$R_{x1}, R_{x2}, \ldots, R_{xJ}$}, Ʒ($R_j$) = ₴($R_j$). / ($\sum_{j=x1}^{xJ}$ ₴($R_j$))

**Proof:** The proof is similar to the proof of property 5. We know from definition 2 that ₴($R_j$) =Ȿ($R_j$)/Ᵽ$_D$. Also, [₴($R_{x1}$)+ ₴($R_{x2}$)+…+ ₴($R_{xJ}$)] = ℘$_{(Vd = v)}$ ($v$ = 0 or 1). So, ₴($R_j$)/[₴($R_{x1}$)+ ₴($R_{x2}$)+…+ ₴($R_{xJ}$)]=Ȿ($R_j$)/℘$_{(Vd=v)}$ = Ʒ($R_j$).

## 5. Sample case study from LBP domain

We illustrate our proposed knowledge representation scheme using a simple example from the LBP domain. The example taken here is not any clinically proved case study; rather a hypothetical situation, drawn on some primary knowledge about LBP, has been depicted for the sake of demonstration. As per our previous assumption, five categories of knowledge sources exist for LBP domain such as national guidelines (level 1), RCT (level 2), existing literature (level 3), patient case studies (level 4), and recommendations of expert physicians (level 5) category.

We consider three symptoms: *LBP without leg pain*, *increased LBP at forward bending*, *LBP aggravated by prolonged sitting*. We define three facts as follows: $f_1$ = ("LBP without leg pain", "yes"), $f_2$ = ("increased LBP at forward bending", "yes"), and $f_3$ = ("LBP aggravated by prolonged sitting", "yes"). Suppose 8 knowledge sources of level 3 category exist which indicate that disease sacroiliac joint pain (SIJ) (Tilvawala et al. 2018; Visser et al. 2013) will be present certainly against $f_1$, 6 knowledge sources of level 5 category are there to indicate that SIJ will be certainly absent for $f_1$, and 7 knowledge sources of level 3 category as well as 5 sources of level 5 category exist to say that SIJ may or may not occur for $f_1$. We also consider that 10 knowledge sources of level 3 category exist to give evidence about the fact that CFJ would certainly occur against the symptom $f_1$, only 1 knowledge source is there which says that CFJ will not occur certainly for $f_1$, and 7 knowledge sources of level 5 category are available which state that CFJ may or may not occur for $f_1$. Discogenic pain may also occur certainly for $f_1$ (supported by 12 knowledge sources of level 3 category). It is also verified by 5 knowledge sources of level 4 category that discogenic pain will never occur for $f_1$. 13 knowledge sources of level 3 category and 7 knowledge sources of level 5 category show that discogenic pain may or may not occur for $f_1$. Whether $f_1$ would be liable for occurrence of disease MPS, is supported by a number of knowledge sources, among which 20 knowledge sources of level 3 category, 15 knowledge sources of level 4 category, and 10 sources of level 5 category claim that MPS would certainly occur for $f_1$. Only 2 knowledge sources of level 5 category state that MPS would certainly be absent from the set of possible diseases for $f_1$. 4 sources of level 4 category and 3 sources of level 5 category support the fact that MPS may or may not occur for $f_1$. To check whether prolapsed inter-vertebral disc (PIVD) (Fairbank (2008)) is there against $f_1$, we get no evidence for certain presence of the disease, but have 16 level 3 sources, 9 level 4 sources, and 11 level 5 sources which say that PIVD would certainly be absent for $f_1$. No evidence is also there to claim both the absence or presence of PIVD for $f_1$.

With a set of knowledge sources for $f_1$ in hand, we now calculate the following using equation (i). Here we use the notations ${T_v}^m(f_1, \text{SIJ})$, ${T_v}^m(f_1, \text{CFJ})$, ${T_v}^m(f_1, \text{DP})$, ${T_v}^m(f_1, \text{MPS})$, and ${T_v}^m(f_1, \text{PIVD})$ (1≤ $m$ ≤3) to denote the acceptability of knowledge sources that find correspondence between the symptom $f_1$ and diseases SIJ, CFJ, discogenic pain, MPS, and PIVD respectively. We show all the calculations in Table 1a and Table 1b. All the calculated values are approximated with a precision of 0.01.

**Table 1a**: Measurement of acceptability of knowledge sources for SIJ, CFJ and discogenic pain against symptom $f_1$

| <$f_1$, SIJ> | <$f_1$, CFJ> | <$f_1$, Discogenic Pain> |
|---|---|---|
| $T_v{}^1(f_1, \text{SIJ}) = \frac{8\times3}{8\times3+6+7\times3+5} \approx 0.43$<br>$T_v{}^2(f_1, \text{SIJ}) = \frac{6}{8\times3+6+7\times3+5} \approx 0.11$<br>$T_v{}^3(f_1, \text{SIJ}) = \frac{7\times3+5}{8\times3+6+7\times3+5} \approx 0.46$ | $T_v{}^1(f_1, \text{CFJ}) = \frac{10\times3}{10\times3+1+7} \approx 0.79$<br>$T_v{}^2(f_1, \text{CFJ}) = \frac{1}{10\times3+1+7} \approx 0.03$<br>$T_v{}^3(f_1, \text{CFJ}) = \frac{7}{10\times3+1+7} \approx 0.18$ | $T_v{}^1(f_1, \text{DP}) = \frac{12\times3}{12\times3+5\times2+13\times3+7} \approx 0.39$<br>$T_v{}^2(f_1, \text{DP}) = \frac{5\times2}{12\times3+5\times2+13\times3+7} \approx 0.11$<br>$T_v{}^3(f_1, \text{DP}) = \frac{13\times3+7}{12\times3+5\times2+13\times3+7} \approx 0.50$ |

**Table 1b**: Measurement of acceptability of knowledge sources for MPS and PIVD against symptom $f_1$

| <$f_1$, MPS> | <$f_1$, PIVD> |
|---|---|
| $T_v{}^1(f_1, \text{MPS}) = \frac{20\times3+15\times2+10}{20\times3+15\times2+10+2+4\times2+3} \approx 0.88$<br>$T_v{}^2(f_1, \text{MPS}) = \frac{2}{20\times3+15\times2+10+2+4\times2+3} \approx 0.02$<br>$T_v{}^3(f_1, \text{MPS}) = \frac{4\times2+3}{20\times3+15\times2+10+2+4\times2+3} \approx 0.10$ | $T_v{}^1(f_1, \text{CFJ}) = \frac{0}{16\times3+9\times2+11} \approx 0$<br>$T_v{}^2(f_1, \text{CFJ}) = \frac{16\times3+9\times2+11}{16\times3+9\times2+11} \approx 1$<br>$T_v{}^3(f_1, \text{CFJ}) = \frac{0}{16\times3+9\times2+11} \approx 0$ |

Now, consider symptom $f_2$. Suppose 9 knowledge sources of level 3 category, 1 source of level 4 category, and 2 sources of level 5 category exist which indicate that disease PIVD will be present certainly against $f_2$, 2 knowledge sources of level 3 category and another 2 of level 4 category are there to indicate that PIVD will be certainly absent for $f_2$, and 5 knowledge sources of level 3 category, 3 sources of level 4 category, as well as 1 source of level 5 category exist to claim that PIVD may or may not occur for $f_2$. We also consider that 17 knowledge sources of level 3 category, 7 source of level 4 category, and 3 sources of level 5 category exist to give evidence about the fact that SIJ would certainly occur against the symptom $f_2$, 11 knowledge sources of level 4 category are there which say that SIJ will not occur certainly for $f_2$, and 9 knowledge sources of level 3 category, 8 sources of level 4 category, and 6 sources of level 5 category are available which state that SIJ may or may not occur for $f_2$. CFJ pain may also occur certainly for $f_2$ (supported by 1 knowledge source of level 3 category and 1 source of level 4 category). As CFJ will be absent certainly for $f_2$ is supported by 2 knowledge sources of level 4 category and 1 source of level 5 category. On the contrary, there are 7 knowledge sources of level 3 category, 10 sources of level 4 category, and 2 sources of level 5 category which provide inconclusive decisions about CFJ for $f_2$.

With a set of knowledge sources for $f_2$ in hand, we now calculate the following using equation (i). Here we use the notations $T_v{}^m(f_2, \text{PIVD})$, $T_v{}^m(f_2, \text{SIJ})$, and $T_v{}^m(f_2, \text{CFJ})$ ($1 \leq m \leq 3$) to denote the acceptability of knowledge sources that find correspondence between the symptom $f_2$ and diseases PIVD, SIJ, and CFJ respectively. We show all the calculations in Table 2a and Table 2b. All the values are approximated with a precision of 0.01.

**Table 2a**: Measurement of acceptability of knowledge sources for PIVD, SIJ against symptom $f_2$

| <$f_2$, PIVD> | <$f_2$, SIJ> |
|---|---|
| $T_v{}^1(f_2, \text{PIVD}) = \frac{9\times3+1\times2+2}{9\times3+1\times2+2+2\times3+2\times2+5\times3+3\times2+1} \approx 0.49$<br>$T_v{}^2(f_2, \text{PIVD}) = \frac{2\times3+2\times2}{9\times3+1\times2+2+2\times3+2\times2+5\times3+3\times2+1} \approx 0.16$<br>$T_v{}^3(f_2, \text{PIVD}) = \frac{5\times3+3\times2+1}{9\times3+1\times2+2+2\times3+2\times2+5\times3+3\times2+1} \approx 0.35$ | $T_v{}^1(f_2, \text{SIJ}) = \frac{17\times3+7\times2+3}{17\times3+7\times2+3+11\times2+9\times3+8\times2+6} \approx 0.49$<br>$T_v{}^2(f_1, \text{CFJ}) = \frac{11\times2}{17\times3+7\times2+3+11\times2+9\times3+8\times2+6} \approx 0.16$<br>$T_v{}^3(f_1, \text{CFJ}) = \frac{9\times3+8\times2+6}{17\times3+7\times2+3+11\times2+9\times3+8\times2+6} \approx 0.35$ |

**Table 2b**: Measurement of acceptability of knowledge sources for CFJ against symptom $f_2$

| <$f_2$, CFJ> |
|---|
| $T_v{}^1(f_2, \text{CFJ}) = \frac{1\times3+1\times2}{1\times3+1\times2+2\times2+1+7\times3+10\times2+2} \approx 0.09$<br>$T_v{}^2(f_2, \text{CFJ}) = \frac{2\times2+1}{1\times3+1\times2+2\times2+1+7\times3+10\times2+2} \approx 0.09$<br>$T_v{}^3(f_2, \text{CFJ}) = \frac{7\times3+10\times2+2}{1\times3+1\times2+2\times2+1+7\times3+10\times2+2} \approx 0.81$ |

Finally, we consider symptom $f_3$. Suppose 6 knowledge sources of level 3 category and 3 sources of level 4 category exist which indicate that disease CFJ will be present certainly against $f_3$, 5 knowledge sources of level 3 category, 2 sources of level 4 category, and 2 sources of level 5 category are there to indicate that CFJ will be certainly absent for $f_3$, and 7 knowledge sources of level 3 category as well as 9 sources of level 4 category and 3 sources of level 5 category exist to say that CFJ may or may not occur for $f_3$. We also consider that 15 knowledge sources of level 3 category, 10 sources of level 4 category, and 11 sources level 5 category exist to give evidence about the fact that PIVD would certainly occur against the symptom $f_3$, 8 knowledge sources of level 3 category, 8 knowledge sources of level 4 category, and 7 sources of level 5 category are there which say that PIVD will not occu$A = \pi r^2$r certainly for $f_3$, and 18 knowledge sources of level 3 category, 12 sources of level 4, and 7 sources of level 5 category are available which state that PIVD may or may not occur for $f_3$. SIJ may also occur certainly for $f_3$ (supported by 7 knowledge sources of level 3 category, 8 sources of level 4 category, and 3 sources of level 5 category). It is also verified by 3 knowledge sources of level 5 category that SIJ will never occur for $f_3$. 17 knowledge sources of level 3 category, 8 sources of level 4 category, and 5 knowledge sources of level 5 category show that SIJ may or may not occur for $f_3$. Whether $f_3$ would be liable for occurrence of discogenic pain, is supported by a number of knowledge sources, among which 15 knowledge sources of level 3 category, 5 knowledge sources of level 4 category, and 6 sources of level 5 category claim that discogenic pain would certainly occur for $f_3$. 6 knowledge sources of level 3 category, another 6 sources of level 4 category, and 3 sources of level 5 category state that discogenic pain would certainly be absent for the set of possible diseases for $f_3$. 7 sources of level 4 category and 4 sources of level 5 category support the fact that discogenic pain may or may not occur for $f_3$.

With a set of knowledge sources for $f_3$ in hand, we now calculate the following using equation (i). Here we use the notations $T_v{}^m(f_3, \text{CFJ})$, $T_v{}^m(f_3, \text{PIVD})$, $T_v{}^m(f_3, \text{SIJ})$, and $T_v{}^m(f_3, \text{DP})$ ($1 \le m \le 3$) to denote the acceptability of knowledge sources that find correspondence between the symptom $f_3$ and diseases CFJ, PIVD, SIJ, and discogenic pain respectively. We show all the calculations in Table 3a and Table 3b. All the calculated values are approximated with a precision of 0.01.

**Table 3a**: Measurement of acceptability of knowledge sources for CFJ, PIVD against symptom $f_3$

| <$f_3$, CFJ> | <$f_3$, PIVD> |
|---|---|
| $T_v{}^1(f_3, \text{CFJ}) = \frac{6 \times 3+3 \times 2}{6 \times 3+3 \times 2+ 5 \times 3+ 2 \times 2+2 + 7 \times 3+9 \times 2+3} \approx 0.28$ <br> $T_v{}^2(f_3, \text{CFJ}) = \frac{5 \times 3+ 2 \times 2+2}{6 \times 3+3 \times 2+ 5 \times 3+ 2 \times 2+2 + 7 \times 3+9 \times 2+3} \approx 0.24$ <br> $T_v{}^3(f_3, \text{CFJ}) = \frac{7 \times 3+9 \times 2+3}{6 \times 3+3 \times 2+ 5 \times 3+ 2 \times 2+2 + 7 \times 3+9 \times 2+3} \approx 0.48$ | $T_v{}^1(f_3, \text{PIVD}) = \frac{15 \times 3+10 \times 2+11}{15 \times 3+10 \times 2+11 +8 \times 3+8 \times 2+7+ 18 \times 3+12 \times 2+7} \approx 0.37$ <br> $T_v{}^2(f_3, \text{PIVD}) = \frac{8 \times 3+8 \times 2+7}{15 \times 3+10 \times 2+11 +8 \times 3+8 \times 2+7+ 18 \times 3+12 \times 2+7} \approx 0.23$ <br> $T_v{}^3(f_3, \text{PIVD}) = \frac{18 \times 3+12 \times 2+7}{15 \times 3+10 \times 2+11 +8 \times 3+8 \times 2+7+ 18 \times 3+12 \times 2+7} \approx 0.41$ |

**Table 3b**: Measurement of acceptability of knowledge sources for SIJ, Discogenic Pain against symptom $f_3$

| <$f_3$, SIJ> | <$f_3$, Discogenic Pain> |
|---|---|
| $T_v{}^1(f_3, \text{SIJ}) = \frac{7 \times 3+8 \times 2+3}{7 \times 3+8 \times 2+3 + 3+ 17 \times 3+8 \times 2+5} \approx 0.35$ <br> $T_v{}^2(f_3, \text{SIJ}) = \frac{3}{7 \times 3+8 \times 2+3 + 3+ 17 \times 3+8 \times 2+5} \approx 0.03$ <br> $T_v{}^3(f_3, \text{SIJ}) = \frac{17 \times 3+8 \times 2+5}{7 \times 3+8 \times 2+3 + 3+ 17 \times 3+8 \times 2+5} \approx 0.63$ | $T_v{}^1(f_3, \text{DP}) = \frac{15 \times 3+5 \times 2+6}{15 \times 3+5 \times 2+6 +6 \times 3+6 \times 2+3+ 7 \times 2+4} \approx 0.54$ <br> $T_v{}^2(f_3, \text{DP}) = \frac{6 \times 3+6 \times 2+3}{15 \times 3+5 \times 2+6 +6 \times 3+6 \times 2+3+ 7 \times 2+4} \approx 0.29$ <br> $T_v{}^3(f_3, \text{DP}) = \frac{7 \times 2+4}{15 \times 3+5 \times 2+6 +6 \times 3+6 \times 2+3+ 7 \times 2+4} \approx 0.16$ |

We now calculate $V_d$ individually for each of the diseases of a symptom. For this, we have to construct a (3×5) matrix for each symptom-disease pair. We denote $K_1$, $K_2$ to represent all the knowledge sources that claim certain presence of a disease corresponding to a symptom, and certain absence of the disease for a symptom respectively. We represent $K_3$ to indicate that the disease may or may not be present for the symptom. The rows hold $K_1$, $K_2$, and $K_3$, while the columns represent level-wise distribution of knowledge sources. Table 4 represents the $M$ matrix (refer to section 4.1) for the pair <$f_1$, SIJ>, denoted by $M(f_1, \text{SIJ})$.

**Table 4**: Design of matrix $M(f_1, \text{SIJ})$ for calculation of $V_d$ for the pair <$f_1$, SIJ>

| | level 1 Category | level 2 Category | level 3 Category | level 4 Category | level 5 Category |
|---|---|---|---|---|---|
| $K_1$ | 0 | 0 | 1 | 0 | 0 |
| $K_2$ | 0 | 0 | 0 | 0 | 1 |
| $K_3$ | 0 | 0 | 1 | 0 | 1 |

As per our strategy for determining $V_d$ for the pair <$f_1$, SIJ> from $M(f_1$, SIJ), we calculate $V_d = 2$. Here we say that the pair <$f_1$, SIJ> takes part in a rule $R_{1a}$. So, $R_{1a}$ is an atomic rule. From the matrix $M(f_1$, SIJ) and the corresponding $T_v{}^m(f_1, \text{SIJ})$, we calculate $CF = 0.46$. We construct similar matrices for the other symptom-disease pairs. We construct a rule $R_{1b}$ to hold <$f_1$, CFJ> with $V_d = 1$, a rule $R_{1c}$ to hold <$f_1$, Discogenic Pain> with $V_d = 2$, a rule $R_{1d}$ to hold <$f_1$, MPS> with $V_d = 1$, a rule $R_{1e}$ to hold <$f_1$, PIVD> with $V_d = 0$. We calculate CF($R_{1b}$) = 0.79, CF($R_{1c}$) = 0.50, CF($R_{1d}$) = 0.88, and CF($R_{1e}$) = 1.00. All the rules are atomic rules.

We follow the same procedure for determining $V_d$ and $CF$ for atomic rules $R_{2a}$ (<$f_2$, PIVD>), $R_{2b}$ (<$f_2$, SIJ>), and $R_{2c}$ (<$f_2$, CFJ>). For $R_{2a}$, $V_d = 1$ and CF($R_{2a}$) = 0.49; for $R_{2b}$, $V_d = 1$ and CF($R_{2b}$) = 0.49; for $R_{2c}$, $V_d = 2$ and CF($R_{2c}$) = 0.81. Also, the same procedure is followed for determining $V_d$ and $CF$ for atomic rules $R_{3a}$ (<$f_3$, CFJ>), $R_{3b}$ (<$f_3$, PIVD>), $R_{3c}$ (<$f_3$, SIJ>), and $R_{3d}$ (<$f_3$, Discogenic Pain>). For $R_{3a}$, $V_d = 2$ and CF($R_{3a}$) = 0.48; for $R_{3b}$, $V_d = 2$ and CF($R_{3b}$) = 0.41; for $R_{3c}$, $V_d = 2$ and CF($R_{3c}$) = 0.63; for $R_{3d}$, $V_d = 1$ and CF($R_{3d}$) = 0.54.

Based on the three facts $f_1$, $f_2$, and $f_3$, we construct a lattice structure that is described in section 4.1. Figure 2 gives the Hasse diagram based on which the data structure (section 4.1.2) is designed.

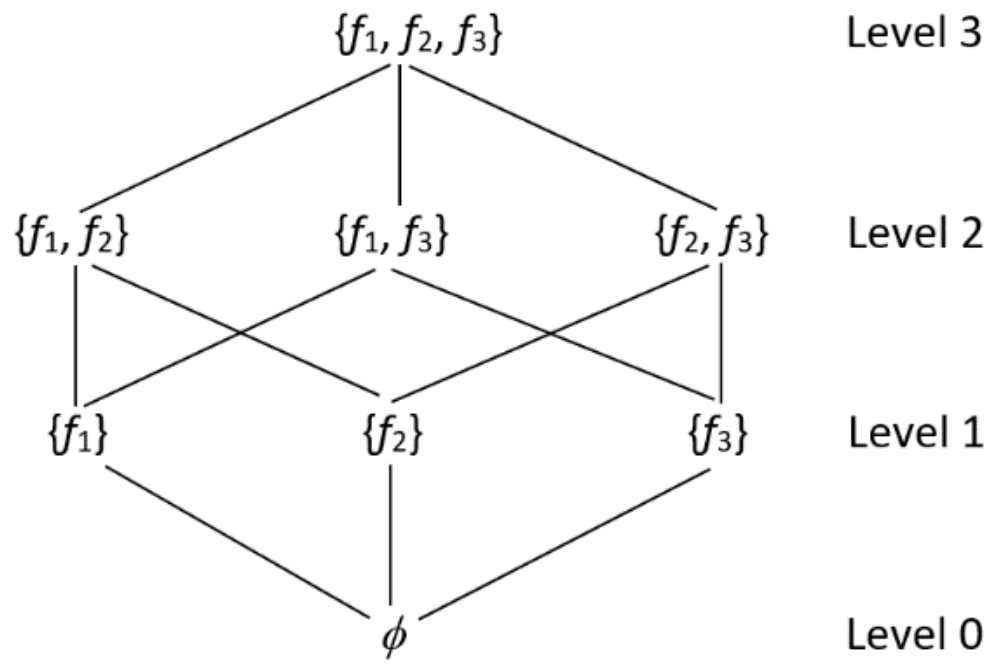

**Figure 2**. Lattice construction based on facts $f_1$, $f_2$, and $f_3$

We construct composite rules for level 2 of the lattice structure depicted in Figure 2. Suppose $R_{12a}$ is a composite rule to hold condition ($f_1$ AND $f_2$) and disease SIJ, with conditional weightages 1/2 and 1/2 for $f_1$ and $f_2$ respectively; $R_{12b}$ is a composite rule to hold condition ($f_1$ AND $f_2$) and disease CFJ, with conditional weightages 2/3 and 1/3 for $f_1$ and $f_2$ respectively; $R_{12c}$ is a composite rule to hold condition ($f_1$ AND $f_2$) and disease discogenic pain, with conditional weightages 2/3 and 1/3 for $f_1$ and $f_2$ respectively; $R_{12d}$ is a composite rule to hold condition ($f_1$ AND $f_2$) and disease MPS, with conditional weightages 2/3 and 1/3 for $f_1$ and $f_2$ respectively; $R_{12e}$ is a composite rule to hold condition ($f_1$ AND $f_2$) and disease PIVD with conditional weightages 1/2 and 1/2 for $f_1$ and $f_2$ respectively. Observing carefully the atomic rules at level 1 of lattice which hold either $f_1$ or $f_2$, we get $V_d = 1$ for $R_{12a}$ and $CF(R_{12a}) = (0.49 \times 0.50 - 0.46 \times 0.50) \approx 0.02$; $V_d = 2$ for $R_{12b}$ and $CF(R_{12b}) = (0.79 \times 2/3 - 0.81 \times 1/3) \approx 0.26$; $V_d = 2$ for $R_{12c}$ and $CF(R_{12c}) = (0.50 \times 2/3) \approx 0.33$; $V_d = 1$ for $R_{12d}$ and $CF(R_{12d}) = (0.88 \times 2/3) \approx 0.59$; $V_d = 0$ for $R_{12e}$ and $CF(R_{12e}) = (1.00 \times 1/2 - 0.49 \times 1/2) \approx 0.26$.

Suppose $R_{13a}$ is another composite rule to hold condition ($f_1$ AND $f_3$) and disease SIJ, with conditional weightages 1/2 and 1/2 for $f_1$ and $f_3$ respectively; $R_{13b}$ is a composite rule to hold condition ($f_1$ AND $f_3$) and disease CFJ, with conditional weightages 1/2 and 1/2 for $f_1$ and $f_3$ respectively; $R_{13c}$ is a composite rule to hold condition ($f_1$ AND $f_3$) and disease discogenic pain, with conditional weightages 2/3 and 1/3 for $f_1$ and $f_3$ respectively; $R_{13d}$ is a composite rule to hold condition ($f_1$ AND $f_3$) and disease PIVD, with conditional weightages 2/3 and 1/3 for $f_1$ and $f_3$ respectively; $R_{13e}$ is a composite rule to hold condition ($f_1$ AND $f_3$) and disease MPS with conditional weightages 2/3 and 1/3 for $f_1$ and $f_3$ respectively. Observing carefully the atomic rules at level 1 of lattice which hold either $f_1$ or $f_3$, we get $V_d = 2$ for $R_{13a}$ and $CF(R_{13a}) = (0.46 \times 0.50 + 0.63 \times 0.50) \approx 0.55$; $V_d = 1$ for $R_{13b}$ and $CF(R_{13b}) = (0.79 \times 0.50 - 0.48 \times 0.50) \approx 0.16$; $V_d = 1$ for $R_{13c}$ and $CF(R_{13c}) = (0.50 \times 2/3 - 0.54 \times 1/3) \approx 0.15$; $V_d = 2$ for $R_{13d}$ and $CF(R_{13d}) = (1 \times 2/3 - 0.41 \times 1/3) \approx 0.53$; $V_d = 1$ for $R_{13e}$ and $CF(R_{13e}) = (0.88 \times 2/3) \approx 0.59$.

Consider another composite rule $R_{23a}$ to hold condition ($f_2$ AND $f_3$) and disease SIJ, with conditional weightages 1/2 and 1/2 for $f_2$ and $f_3$ respectively; $R_{23b}$ is a composite rule to hold condition ($f_2$ AND $f_3$) and disease CFJ, with conditional weightages 1/2 and 1/2 for $f_2$ and $f_3$ respectively; $R_{23c}$ is a composite rule to hold condition ($f_2$ AND $f_3$) and disease PIVD, with conditional weightages 1/2 and 1/2 for $f_2$ and $f_3$ respectively; $R_{23d}$ is a composite rule to hold condition ($f_2$ AND $f_3$) and disease discogenic pain, with conditional weightages 1/2 and 1/2 for $f_2$ and $f_3$ respectively; Observing carefully the atomic rules at level 1 of lattice which hold either $f_2$ or $f_3$, we get $V_d = 2$ for $R_{23a}$ and $CF(R_{23a}) = (0.63 \times 0.50 - 0.49 \times 0.50) \approx 0.07$; $V_d = 2$ for $R_{23b}$ and $CF(R_{23b}) = (0.81 \times 0.50 + 0.48 \times 0.50) \approx 0.65$; $V_d = 1$ for $R_{23c}$ and $CF(R_{23c}) = (0.49 \times 0.50 - 0.41 \times 0.50) \approx 0.04$; $V_d = 1$ for $R_{23d}$ and $CF(R_{23d}) = (0.54 \times 0.50) \approx 0.27$.

Now we construct a composite rule $R_{123a}$ to hold condition ($f_1$ AND $f_2$ AND $f_3$) and disease SIJ, with conditional weightages 1/3, 1/3, and 1/3 for $f_1$, $f_2$, and $f_3$ respectively; $R_{123b}$ is a composite rule to hold condition ($f_1$ AND $f_2$ AND $f_3$) and disease CFJ, with conditional weightages 1/3, 1/3, and 1/3 for $f_1$, $f_2$, and $f_3$ respectively; $R_{123c}$ is a composite rule to hold condition ($f_1$ AND $f_2$ AND $f_3$) and disease PIVD, with conditional weightages 2/4, 1/4, and 1/4 for $f_1$, $f_2$, and $f_3$ respectively; $R_{123d}$ is a composite rule to hold condition ($f_1$ AND $f_2$ AND $f_3$) and disease discogenic pain, with conditional weightages 2/5, 1/5, and 2/5 for $f_1$, $f_2$, and $f_3$ respectively; $R_{123e}$ is a composite rule to hold condition ($f_1$ AND $f_2$ AND $f_3$) and disease MPS, with conditional weightages 1/3, 1/3, and 1/3 for $f_1$, $f_2$, and $f_3$ respectively. Observing carefully all the composite rules at level 2 of lattice, we get $V_d = 2$ for $R_{123a}$ and $CF(R_{123a}) = 1/2 \times [(0.55 - 0.02)\times1/3 + (0.07 - 0.02)\times1/3 + (0.55 + 0.07)\times1/3] \approx 0.21$; $V_d = 2$ for $R_{123b}$ and $CF(R_{123b}) = 1/2 \times [(0.26 - 0.16)\times1/3 + (0.26 + 0.65)\times1/3 + (0.65 - 0.16)\times1/3] \approx 0.25$; $V_d = 2$ for $R_{123c}$ and $CF(R_{123c}) = 1/2 \times [(0.53 - 0.26)\times2/4 + (0.53 - 0.04)\times1/4 + (0.26 - 0.04)\times1/4] \approx 0.16$; $V_d = 2$ for $R_{123d}$ and $CF(R_{123d}) = 1/2 \times [(0.33 - 0.15)\times2/5 + (0.15 + 0.27)\times2/5 + (0.33 - 0.27)\times1/5] \approx 0.13$; $V_d = 1$ for $R_{123e}$ and $CF(R_{123e}) = 1/2 \times [(0.59 + 0.59)\times1/3 + 0.59 \times 1/3 + 0.59 \times 1/3] \approx 0.40$.

So, now we have a set of 31 un-optimized rules, some of which also contain inconsistent knowledge. We would use rough set theory to generate optimal rules after handling the inconsistencies in knowledge. In this example, we have five LBP diseases namely SIJ, CFJ, PIVD, discogenic pain, and MPS. For each of diseases we will find two concepts as described in section 4.2.1, which are given below:

Concept$_1$(SIJ) = $\{R_{1a}, R_{2b}, R_{3c}, R_{12a}, R_{13a}, R_{23a}, R_{123a}\}$, Concept$_2$(SIJ) = $\{R_{1a}, R_{3c}, R_{13a}, R_{23a}, R_{123a}\}$
Concept$_1$(CFJ) = $\{R_{1b}, R_{2c}, R_{3a}, R_{12b}, R_{13b}, R_{23b}, R_{123b}\}$, Concept$_2$(CFJ) = $\{R_{2c}, R_{3a}, R_{12b}, R_{23b}, R_{123b}\}$
Concept$_1$(PIVD) = $\{R_{2a}, R_{3b}, R_{13d}, R_{23c}, R_{123c}\}$, Concept$_2$(PIVD) = $\{R_{1e}, R_{3b}, R_{12e}, R_{13d}, R_{123c}\}$
Concept$_1$(Discogenic pain) = $\{R_{1c}, R_{3d}, R_{12c}, R_{13c}, R_{23d}, R_{123d}\}$, Concept$_2$(Discogenic pain) = $\{R_{1c}, R_{12c}, R_{123d}\}$
Concept$_1$(MPS) = $\{R_{1d}, R_{12d}, R_{13e}, R_{123e}\}$, Concept$_2$(MPS) = $\phi$

As, Concept$_1$(SIJ) ∩ Concept$_2$(SIJ) ≠ $\phi$, we have to determine the lower and upper approximations of both the concepts. Similarly, for Concept$_1$(CFJ) and Concept$_2$(CFJ), Concept$_1$(PIVD) and Concept$_2$(PIVD), Concept$_1$(Discogenic pain) and Concept$_2$(Discogenic pain), we have to determine the lower and upper approximations as per Approximation_Generation_KB procedure.

Lower approximation of Concept$_1$(SIJ) = $\{R_{2b}, R_{12a}\}$
Upper approximation of Concept$_1$(SIJ) = $\{R_{1a}, R_{2b}, R_{3c}, R_{12a}, R_{13a}, R_{23a}, R_{123a}\}$
Boundary region of Concept$_1$(SIJ) = $\{R_{1a}, R_{3c}, R_{13a}, R_{23a}, R_{123a}\}$
Lower approximation of Concept$_2$(SIJ) = $\phi$
Upper approximation of Concept$_2$(SIJ) = $\{R_{1a}, R_{3c}, R_{13a}, R_{23a}, R_{123a}\}$
Boundary region of Concept$_2$(SIJ) = $\{R_{1a}, R_{3c}, R_{13a}, R_{23a}, R_{123a}\}$

Lower approximation of Concept$_1$(CFJ) = $\{R_{1b}, R_{13b}\}$
Upper approximation of Concept$_1$(CFJ) = $\{R_{1b}, R_{2c}, R_{3a}, R_{12b}, R_{13b}, R_{23b}, R_{123b}\}$
Boundary region of Concept$_1$(CFJ) = $\{R_{2c}, R_{3a}, R_{12b}, R_{23b}, R_{123b}\}$
Lower approximation of Concept$_2$(CFJ) = $\phi$
Upper approximation of Concept$_2$(CFJ) = $\{R_{2c}, R_{3a}, R_{12b}, R_{23b}, R_{123b}\}$
Boundary region of Concept$_2$(CFJ) = $\{R_{2c}, R_{3a}, R_{12b}, R_{23b}, R_{123b}\}$

Lower approximation of Concept$_1$(PIVD) = $\{R_{2a}, R_{23c}\}$

Upper approximation of $Concept_1$(PIVD) = {$R_{2a}$, $R_{3b}$, $R_{13d}$, $R_{23c}$, $R_{123c}$}
Boundary region of $Concept_1$(PIVD) = {$R_{3b}$, $R_{13d}$, $R_{123c}$}
Lower approximation of $Concept_2$(PIVD) = {$R_{1e}$, $R_{12e}$}
Upper approximation of $Concept_2$(PIVD) = {$R_{1e}$, $R_{3b}$, $R_{12e}$, $R_{13d}$, $R_{123c}$}
Boundary region of $Concept_2$(PIVD) = {$R_{3b}$, $R_{13d}$, $R_{123c}$}

Lower approximation of $Concept_1$(Discogenic pain) = {$R_{3d}$, $R_{13c}$, $R_{23d}$}
Upper approximation of $Concept_1$(Discogenic pain) = {$R_{1c}$, $R_{3d}$, $R_{12c}$, $R_{13c}$, $R_{23d}$, $R_{123d}$}
Boundary region of $Concept_1$(Discogenic pain) = {$R_{1c}$, $R_{12c}$, $R_{123d}$}
Lower approximation of $Concept_2$(Discogenic pain) = $\phi$
Upper approximation of $Concept_2$(Discogenic pain) = {$R_{1c}$, $R_{12c}$, $R_{123d}$}
Boundary region of $Concept_2$(Discogenic pain) = {$R_{1c}$, $R_{12c}$, $R_{123d}$}

Now, we generate certain and optimal decision rules using the lower approximations of concepts and applying the rule-minimization technique.

Rule 1. (“LBP aggravated by prolonged sitting”, “no”) AND (“increased LBP at forward bending”, “yes”) $\rightarrow$ (SIJ,1) with reliability strength = 0.12.

Rule 2. (“LBP without leg pain”, “yes”) AND (“increased LBP at forward bending”, “no”) $\rightarrow$ (CFJ, 1) with reliability strength = 0.16.

Rule 3. (“LBP without leg pain”, “no”) AND (“increased LBP at forward bending”, “yes”) $\rightarrow$ (PIVD, 1) with reliability strength = 0.20.

Rule 4. (“LBP without leg pain”, “yes”) AND (“LBP aggravated by prolonged sitting”, “no”) $\rightarrow$ (PIVD, 0) with reliability strength = 0.29.

Rule 5. (“Increased LBP at forward bending”, “no”) AND (“LBP aggravated by prolonged sitting”, “yes”) $\rightarrow$ (Discogenic pain, 1) with reliability strength = 0.24.

Rule 6. (“LBP without leg pain”, “no”) AND (“LBP aggravated by prolonged sitting”, “yes”) $\rightarrow$ (Discogenic pain,1) with reliability strength = 0.28.

Rule 5 and Rule 6 can be further minimized as: (“LBP aggravated by prolonged sitting”, “yes”) AND ((“LBP without leg pain”, “no”) OR (“Increased LBP at forward bending”, “no”)) $\rightarrow$ (Discogenic pain, 1)

Rule 7. (“LBP without leg pain”, “yes”) $\rightarrow$ (MPS, 1) with reliability strength = 1.00.

Here, we do not show the uncertain optimal rules as a clinical diagnosis should not allow, in most cases, the uncertain rules. The certain rules for a particular disease should be arranged in descending order according to their reliability strengths, i.e., the rule whose strength is high would be applicable first in case of backward reasoning. If no diagnostic decision is taken based on the certain rules, only then the uncertain rules are taken into account after consulting with domain experts.

## 6. Discussion of properties of proposed knowledge representation

Our proposed lattice-based knowledge representation scheme follows a number of properties as follows.

*Completeness*
Using our proposed knowledge representation, we can design a complete knowledge base KB as, based on *n* atomic rules, we can generate an exhaustive set of ($2^n - n - 1$) composite rules. No other rules are possible from the set of atomic rules present in a knowledge module.

*Consistency*

The proposed scheme is consistent as we have generated the lower and upper approximations of concepts for each inconsistent decision. Also, in case of modification of the rules available in KB, we have kept provisions for generations of approximated sets if any kind of inconsistency is seen in the modified KB.

*Integriety*
The use of lattice structure for the design of KB allows to support integriety among the related rules in a knowledge module. As we have used the concept of '*glb*' and '*lub*' of a lattice, the condition part and decision part of a rule are designed based on the conditions and decisions of corresponding predecessor rules. Modification in KB also is performed based on the predecessor and successor nodes of a particular *Lattice_Element* node. So, the KB works in an integrated manner.

*Zero Redundancy*
The proposed design does not support redundant information, as we have incorporated lattice theory in our structure. As lattice theory does not allow duplicate set elements, there are no redundant rules in the KB. Also, applying the rule minimization technique on the decision rules generated from the approximated sets of concepts, we have removed any further redundancy that may be hidden within the condition parts of decision rules.

*Ease of Access*
As we have proposed a level-wise design of KB, rules with $k$ facts ($1 \leq k \leq n$) in condition parts appear at $(k-1)^{th}$ level of lattice. So, if we want to find a rule with $k$ facts, we would directly access the $(k-1)^{th}$ level. Also, from the lattice structure we can find easily those rules which include fact $f$ in their condition parts. Furthermore, the unique labels assigned to the *Lattice_Element* nodes are used for their easy identification. As a result, the access to the *Lattice_Element* nodes becomes quicker.

## 7. Conclusion

In this paper, we have indicated how rough set based lattice structure for knowledge representation provides foundation for designing a knowledge base for a medical expert system. As a medical expert system should be reliable, we have proposed a new metric for a rule to measure how much trustworthy the rule is. The proposed knowledge representation scheme generates optimal decision rules which may be certain, possible or uncertain. But as reliability is a major concern in medical expert systems, these kind of systems would mainly consider the certain rules. We also have derived some probabilistic properties of the optimal rules. We have illustrated the whole approach using a simple case study from the domain of low back pain. The proposed design has the properties like completeness, consistency, integrity, and non-redundancy. The design also facilitates ease of access of the rules. We plan to use this scheme for the real-world development of the knowledge base for a medical expert system for LBP management in the near future.

## Appendices

### Appendix A

/*The procedure *Create_Level_Node* is described below*/

Procedure *Create_Level_Node* (*prev*, *level*) **/*** *prev* is the *Lattice_Level* node at the immediate predecessor level and *level* is the current level at which new *Lattice_Level* node is to be created */

**/*** Allocating space for a Lattice_Level node */

$l_v$ = *Get_Level_Node*();

**/***Assigning values to the elements belonging to the structure of level node*/

*next_level* [*prev*] = $l_v$;

*next_level* [$l_v$] = NIL;

*level* [$l_v$] = *level*;

*total_node_level* [$l_v$] = $^{n}C_{level}$;

*arr* [$l_v$] = Size_of ((Lattice_Element node) * *total_node_level* [$l_v$]) /* Allocation of total space for the array of $l_v$ */

For *i* = 0 to *total_node_level* [$l_v$]–1 Do

*arr*[$l_v$]$_i$ = NIL;

Endfor

**/*** Returning the address of the newly created level node to the *Build_KB* procedure */

Return $l_v$ ;

End *Create_Level_Node*

/* The procedure *Create_Lattice_Node* is described below. */

Precedure *Create_Lattice_Node* ($l_v$) /* $l_v$ is the *Lattice_Level* node where a new *Lattice_Element* node is to be created */

/* Allocating space for a *Lattice_Element* node */

$l_n$ = *Get_Element_Node*();

/*Assigning values to the structural components of a *Lattice_Element* node */

*label* [$l_n$] = *Assign_Label* ($l_v$, $l_n$); /* The procedure *Assign_Label* assigns a unique label to $l_n$ */

*condition* [$l_n$] = *Assign_Condition* ($l_n$, $l_v$); /**Assign_Condition* procedure assigns condition to $l_n$ */

*decision* [$l_n$] = *Assign_Decision* ($l_n$, $l_v$); /* *Assign_Decision* procedure assigns decision to $l_n$ */

*predecessor* [$l_n$] = NIL;

*successor* [$l_n$] = NIL;

**/*** Returning the address of the newly created *Lattice_Element* node to the *Build_KB* procedure */

Return $l_n$ ;

End *Create_Lattice_Node*

/* The procedure *Create_Predecessor_List* is described below. */

Procedure *Create_Predecessor_List* ($l_n$, *level*) /* $l_n$ is a *Lattice_Element* node and *level* represents the current lattice level */

/* Initialization of pointers */

*h* = NIL , *temp* = NIL, $p_n$ =NIL; /* *h* is the starting node of the predecessor list of $l_n$, *temp* is a temporary *Node* pointer, and $p_n$ is a *Node* pointer to hold newly generated node. */

/* Space allocation for the initial node of predecessor list of $l_n$ */

If *level* == 0 Then

*h* = NIL;

Else

*h* = *Get_Node*(); //The procedure allocates space for the *h* node

*label*[*h*] = $0^n$ ;

*ptr*[*h*] = NIL;

Endif

/* Generation of consecutive predecessor nodes */

*temp* = *h*

For *i* = 1 to *level* Do

$p_n$ = *Get_Node*;

*label*[$p_n$] = *Predecessor_Label*(*label*[$l_n$], *i*);

*ptr*[$p_n$] = NIL;

*ptr*[*temp*] = $p_n$;

*temp* = *ptr*[*temp*];

```
        Endfor
        /* Returning the address of the first node of the predecessor list of l_n to the Build_KB procedure */
        Return h;
End Create_Predecessor_List
```

The procedure *Predecessor_Label* (*label, j*) searches for *j*-th 1 in the bit string of '*label*' starting from the leftmost position. Suppose the location of *j*-th 1 in the bit string is *k*. The procedure assigns 0 at the *k*-th position of the bit string and returns the modified bit string.

```
/* The procedure Create_Successor_List is described below. */

Procedure Create_Successor_List(l_n, level)  /* l_n is a Lattice_Element node and level represents the current lattice
                                                level */
        /* Initialization of pointers */
        h = NIL , temp = NIL, s_n = NIL;  /* h is the starting node of the predecessor list of l_n, temp is a temporary
                                           Node pointer, and s_n is a Node pointer to hold newly generated node. */
        /* Space allocation for the initial node of successor list of l_n */
       If level == n Do
             h = NIL;
       Endif
       /* Generation of consecutive successor nodes */
       If level < n Then
               h = Get_Node();
               label[h] = NIL;
               ptr[h] = NIL;
               temp = h;
               For i = 1 to (n – level)
                      s_n = Get_Node ();
                      label[s_n] = Successor_Label(label[l_n], i);
                      ptr[s_n] = NIL;
                      ptr[temp] = s_n;
                      temp = ptr[temp];
               Endfor
        Endif
        /* Returning the address of the first node of the successor list of l_n to the Build_KB procedure */
        Return h;
End Create_Successor_List
```

The procedure *Successor_Label*(*label*, *j*) searches for *j*-th 0 in the bit string of '*label*' starting from the rightmost position. Suppose the location of *j*-th 0 in the bit string is *k*. The procedure assigns 1 at the *k*-th position of the bit string and returns the modified bit string.

**Appendix B**

/*The following procedure is for approximation generation from modified knowledge when decision *d* is also included in some other *Lattice_Element* nodes in KB*/

```
Procedure Modification_Approximation_Generation_1 (ArrD, NewD) /* NewD is the set of Lattice_Element nodes
                                                        which now include d in their decision parts */
       /* Initialization of variables */
       count = Size_of(NewD);                  //count is a temporary variable to hold the size of NewD
       /* Generation of modified approximated sets for Concept_1. Note that Concept_1 has been modified accordingly */
       l_1 = |Upper_Approximation_1|, l_2 = |Lower_Approximation_1|, l_3 = |Boundary_Region_1|;
       For i = 0 to count – 1Do
            If V_d == 1 OR V_d == 2 Then
                 Upper_Approximation_1[l_1] = Upper_Approximation_1 ∪ {NewD_i};
```

$l_1 = l_1 + 1$;
Endif
If $V_d == 1$ Then
*Lower_Approximation*$_1[l_2]$ = *Lower_Approximation*$_1 \cup \{NewD_i\}$;
$l_2 \leftarrow l_2 + 1$;
Endif
If $V_d == 2$ Then
*Boundary_Region*$_1[l_3]$ = *Boundary_Region*$_1 \cup \{NewD_i\}$;
$l_3 = l_3 + 1$;
Endif
Endfor
/* Generation of modified approximated sets for Concept$_2$. Note that Concept$_2$ has been modified accordingly ***/**
$l_1 = |$*Upper_Approximation*$_2|$, $l_2 = |$*Lower_Approximation*$_2|$, $l_3 = |$*Boundary_Region*$_2|$;
For $i = 0$ to *count* – 1 Do
If $V_d == 0$ OR $V_d == 2$ Then
*Upper_Approximation*$_2[l_1]$ = *Upper_Approximation*$_2 \cup \{NewD_i\}$;
$l_1 = l_1 + 1$;
Endif
If $V_d == 0$ Then
*Lower_Approximation*$_2[l_2]$ = *Lower_Approximation*$_2 \cup \{NewD_i\}$;
$l_2 = l_2 + 1$;
Endif
If $V_d == 2$ Then
*Boundary_Region*$_2[l_3]$ = *Boundary_Region*$_2 \cup \{NewD_i\}$;
$l_3 = l_3 + 1$;
Endif
Endfor
End *Modification_Approximation_Generation*$_1$

/* The following procedure is for approximation generation from modified knowledge when decision *d* is deleted from some existing *Lattice_Element* nodes in KB. */

Procedure *Modification_Approximation_Generation*$_2$ (*ArrD*, *OldD*) /* *OldD* is the set of *Lattice_Element* nodes from which decisions having disease *d* are to be deleted */
**/*** Initialization of variables */
*count* = Size_of(*OldD*); //*count* is a temporary variable to hold the size of *OldD*
/* Generation of modified approximated sets for Concept$_1$. Note that Concept$_1$ has been modified accordingly ***/**
$l_1 = |$*Upper_Approximation*$_1|$, $l_2 = |$*Lower_Approximation*$_1|$, $l_3 = |$*Boundary_Region*$_1|$;
For $i = 0$ to *count* – 1 Do
If $V_d == 1$ OR $V_d == 2$ Then
*Upper_Approximation*$_1[l_1]$ = *Upper_Approximation*$_1 - \{OldD_i\}$;
$l_1 \leftarrow l_1 - 1$;
Endif
If $V_d == 1$ Then
*Lower_Approximation*$_1[l_2]$ = *Lower_Approximation*$_1 - \{OldD_i\}$;
$l_2 = l_2 - 1$;
Endif
If $V_d == 2$ Then
*Boundary_Region*$_1[l_3]$ = *Boundary_Region*$_1 - \{OldD_i\}$;
$l_3 = l_3 - 1$;
Endif
Endfor
/* Generation of modified approximated sets for Concept$_2$. Note that Concept$_2$ has been modified accordingly ***/**
$l_1 = |$*Upper_Approximation*$_2|$, $l_2 = |$*Lower_Approximation*$_2|$, $l_3 = |$*Boundary_Region*$_2|$;
For $i = 0$ to *count* – 1 Do

```
            If Vd == 0 OR Vd == 2 Then
                Upper_Approximation2[l1] = Upper_Approximation2 – {OldDi};
                l1 ← l1 – 1;
            Endif
            If Vd == 0 Then
                Lower_Approximation2[l2] = Lower_Approximation2 – {OldDi};
                l2 = l2 – 1;
            Endif
            If Vd == 2 Then
                Boundary_Region2 [l3] = Boundary_Region2 – {OldDi};
                l3 = l3 – 1;
            Endif
        Endfor
End Modification_Approximation_Generation2
```

/* The following procedure is for approximation generation from modified knowledge when $V_d$ for decision *d* in a particular *Lattice_Element* node in KB has been changed. */

```
Procedure Modification_Approximation_Generation3 (ln, d)   /*ln is the Lattice_Element node whose Vd
                                                              has been changed to Vd_new */
        If Vd == 0 AND Vd_new == 1 Then
            Upper_Approximation1 = Upper_Approximation1 ∪ {ln};
            Lower_Approximation1 = Lower_Approximation1 ∪ {ln};
            Boundary_Region1 = Boundary_Region1 ∪ {ln};
            Upper_Approximation2 = Upper_Approximation2 – {ln};
            Lower_Approximation2 = Lower_Approximation2 – {ln};
            Boundary_Region2 = Boundary_Region2 – {ln};
        Elseif Vd == 0 AND Vd_new == 2 Then
            Upper_Approximation1 = Upper_Approximation1 ∪ {ln};
            Boundary_Region1 = Boundary_Region1 ∪ {ln};
            Upper_Approximation2 = Upper_Approximation2 – {ln};
            Lower_Approximation2 = Lower_Approximation2 – {ln};
            Boundary_Region2 = Boundary_Region2 – {ln};
        Elseif Vd == 1 AND Vd_new == 0 Then
            Upper_Approximation1 = Upper_Approximation1 – {ln};
            Lower_Approximation1 = Lower_Approximation1 – {ln};
            Boundary_Region1 = Boundary_Region1 – {ln};
            Upper_Approximation2 = Upper_Approximation2 ∪ {ln};
            Lower_Approximation2 = Lower_Approximation2 ∪ {ln};
            Boundary_Region2 = Boundary_Region2 ∪ {ln};
        Elseif Vd == 1 AND Vd_new == 2 Then
            Lower_Approximation1 = Lower_Approximation1 – {ln};
            Boundary_Region1 = Boundary_Region1 ∪ {ln};
        Elseif Vd == 2 AND Vd_new == 0 Then
            Upper_Approximation1 = Upper_Approximation1 – {ln};
            Boundary_Region1 = Boundary_Region1 – {ln};
            Upper_Approximation2 = Upper_Approximation2 ∪ {ln};
            Lower_Approximation2 = Lower_Approximation2 ∪ {ln};
            Boundary_Region2 = Boundary_Region2 ∪ {ln};
        Elseif Vd == 2 AND Vd_new == 1 Then
            Lower_Approximation1 = Lower_Approximation1 ∪ {ln};
            Boundary_Region1 = Boundary_Region1 – {ln};
        Endif
End Modification_Approximation_Generation3
```